\documentclass{MITcsail}

\usepackage{amsmath,amsfonts,bm}

\def\eqref#1{equation~\ref{#1}}

\def\1{\bm{1}}

\DeclareMathAlphabet{\mathsfit}{\encodingdefault}{\sfdefault}{m}{sl}
\SetMathAlphabet{\mathsfit}{bold}{\encodingdefault}{\sfdefault}{bx}{n}

\definecolor{darkred}{RGB}{140, 21, 21}
\definecolor{lightgray}{gray}{0.5}
\definecolor{orange}{HTML}{F58025}
\definecolor{deepred}{rgb}{0.631,0.102,0.102}
\definecolor{amethyst}{rgb}{0.6, 0.4, 0.8}
\definecolor{darkgreen}{rgb}{0.3,0.7,0.3}
\definecolor{salmon}{RGB}{241, 150, 141}
\definecolor{mildyellow}{HTML}{FFF2CC}
\definecolor{jhulightblue}{HTML}{68ACE5}

\usepackage{hyperref}
\usepackage{xcolor}
\hypersetup{
    colorlinks=true,
    linkcolor=jhulightblue,
    citecolor=lightgray,
    filecolor=darkred,
    urlcolor=jhulightblue
}
\usepackage[authoryear]{natbib}
\usepackage{amsmath}
\usepackage{amsfonts}
\usepackage{amssymb}
\usepackage{wrapfig}
\usepackage{subcaption}
\usepackage{multirow}
 \usepackage{mathtools}

\usepackage{verbatim}

\usepackage{anyfontsize}

\usepackage{microtype}
\usepackage{graphicx}
\usepackage{booktabs}
\usepackage{multirow}
\usepackage{amsmath,amssymb}
\usepackage{booktabs}
\usepackage{caption,subcaption}
\usepackage{svg}

\usepackage{authblk}

\renewcommand\Affilfont{\normalsize\normalfont}

\definecolor{mygreen}{HTML}{3cb44b}
\definecolor{skyblue}{HTML}{beffff}
\definecolor{lightgreen}{HTML}{90ee90}

\usepackage{color, colortbl}

\definecolor{emerald}{rgb}{0.31, 0.78, 0.37}

\usepackage{datetime}
\newdateformat{ymd}{\the\year-\the\month-\the\day}

\usepackage{tcolorbox}
\usepackage{enumitem}
\definecolor{mygreen}{HTML}{3cb44b}
\colorlet{myyellow}{green!10!orange!90!}
\makeatletter

\usepackage{tikz}
\usetikzlibrary{arrows,shapes,snakes,automata,backgrounds,fit,petri}
\usepackage{adjustbox}

\newcommand{\RN}[1]{\textup{\lowercase\expandafter{\it \romannumeral#1}}}
\usepackage{tabu}

\def\KL{\textsf{KL}}

\newcommand{\beq}{\vspace{0mm}\begin{equation}}
\newcommand{\eeq}{\vspace{0mm}\end{equation}}
\newcommand{\beqs}{\vspace{0mm}\begin{eqnarray}}
\newcommand{\eeqs}{\vspace{0mm}\end{eqnarray}}
\newcommand{\barr}{\begin{array}}
\newcommand{\earr}{\end{array}}

\usepackage{color, colortbl}
\definecolor{Gray}{gray}{0.93}

\usepackage{lipsum}

\usepackage{pifont}

\usepackage{makecell}

\usepackage{xcolor,amsmath}
\usepackage[ruled,vlined,linesnumbered]{algorithm2e}
\SetKwComment{Comment}{$\triangleright$ }{}
\SetKwInput{KwInput}{Input}
\SetKwInput{KwOutput}{Output}
\DontPrintSemicolon

\usepackage{xcolor}
\definecolor{mygreen}{HTML}{3cb44b}

\SetKwComment{Comment}{\color{green!50!black}\# }{}

\SetKwProg{Function}{def}{:}{}

\SetKwProg{For}{for}{:}{}
\SetKwProg{If}{if}{:}{}

\usepackage[T1]{fontenc}
\usepackage{listings}
\usepackage{float}
\usepackage{array}
\usepackage{url}

\newcommand{\pagehead}{}

\newcommand{\symbolqa}{\textsc{Symbol-QA}}
\newcommand{\llmqa}{\textsc{LLM-QA}}
\newcommand{\realqa}{\textsc{Real-QA}}
\newcommand{\finalacc}{\mathrm{Final}}
\newcommand{\diagacc}{\mathrm{Diag}}
\newcommand{\forget}{\mathrm{Forget}}
\renewcommand{\KL}{\mathrm{KL}}

\usepackage{soul}

\title{Continual Learning Mechanisms Compose for Long-Horizon Memorization}

\author
{Zheyuan Zhang\textsuperscript{$*$}, Alvin Zhang\textsuperscript{$*$}, Daniel Khashabi\textsuperscript{$\dagger$}, Tianmin Shu\textsuperscript{$\dagger$} \\
\vspace{0.6em}
{\normalfont\sffamily Johns Hopkins University} \\
\vspace{0.6em}
\texttt{\href{https://compose-cl.github.io/}{compose-cl.github.io}}
}
\begin{document}

\maketitle

\thispagestyle{firstpagestyle}
\renewcommand\thefootnote{}\footnote{$^{*}$ Equal contribution. $^{\dagger}$ Equal advising.}

\paragraph{\textit{Abstract.}}
Language models may need to internalize information that arrives over time and retain it through many subsequent updates. To study this challenge, we introduce long-horizon \emph{memorization}, a setting in which a model learns 100 query-answer tasks through continual supervised fine-tuning without retaining earlier training examples or receiving task identifiers at inference. Sequential updates cause catastrophic forgetting, and no single continual learning mechanism we evaluate maintains strong retention at this horizon. We hypothesize that mechanisms addressing complementary sources of forgetting will be more effective when composed. We organize these compositions along two design dimensions. Data, function, and weight anchors specify what prior information each update should preserve, while low-rank allocation rules determine where successive updates are retained. To test this hypothesis systematically, we construct three distinct 100-task memorization datasets. We introduce task-level successive halving to search the combinatorial design space and use a factorial experiment to measure individual and interaction effects. Our best method combines all three anchors with merged LoRA, ranks among the top 3 methods in all datasets, and raises average final retention from 1.2\% under naive sequential fine-tuning to 34.9\%, a \textbf{28-fold} improvement. The data anchor and merged LoRA provide the largest average gains and interact super-additively on all three datasets. Together, these results show that composing complementary mechanisms substantially improves long-horizon memorization beyond what any individual mechanism achieves.

\section{Introduction}

\begin{figure}[!t]
\centering
\includegraphics[width=\linewidth]{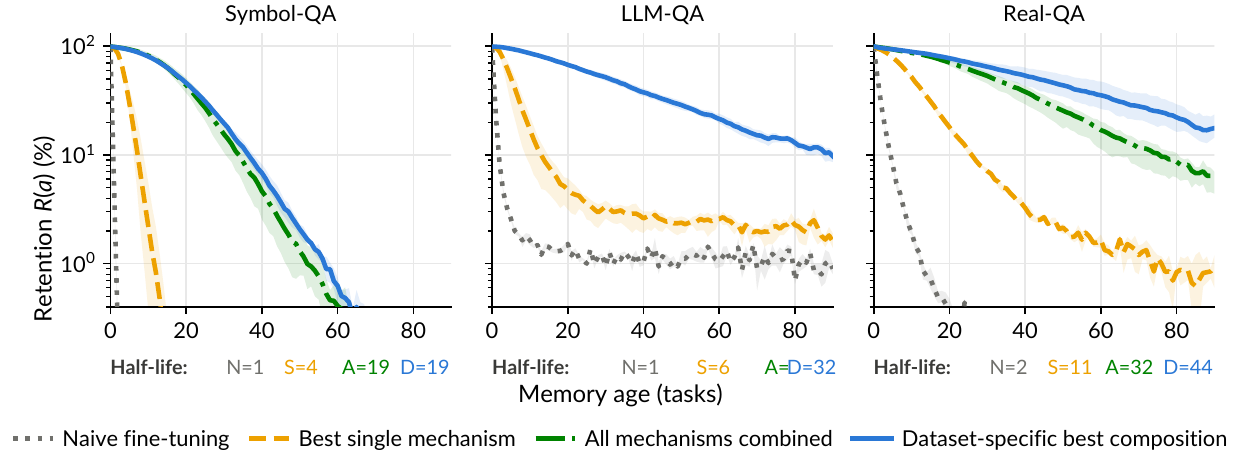}
\caption{Memory lifetime under the best single mechanism and composed continual learning methods. Lines show three-seed means with min-max bands. Half-life counts tasks until retention halves. Combining multiple anchors with merged LoRA substantially reduces catastrophic forgetting and extends memory lifetime beyond the best single mechanism.
}
\label{fig:survival}
\end{figure}

Consider a language model that learns new information over time by updating its parameters. For these parameters to serve as memory, the model must remember what it has learned even after many more updates. We call this setting long-horizon \emph{memorization}. Prompting and retrieval can provide new information at inference time \citep{brown2020language,lewis2020retrieval}, but the information remains outside the model's parameters and must be supplied again. We instead ask whether repeated updates can build and preserve this memory within the model itself.

We study this problem through continual supervised fine-tuning (SFT) in the domain-incremental setting \citep{van2019three}. Each task contains a set of query-answer pairs. The model learns 100 tasks in sequence without retaining raw examples from earlier tasks, and it receives no task identifier at inference. The goal is to learn each new task while retaining associations learned from previous tasks. This is difficult because updates for new tasks can overwrite previously stored knowledge, causing catastrophic forgetting \citep{mccloskey1989catastrophic,french1999catastrophic}.

Prior work shows that mixing rehearsal with knowledge distillation has strong performance \citep{buzzega2020dark}, but it does not systematically study the broader space of mechanism compositions. We therefore hypothesize that mechanisms addressing complementary sources of forgetting will retain associations more effectively when composed. To test this hypothesis, we organize compositions along two design dimensions: anchors and low-rank allocation rules. Anchors specify what prior information an update should preserve. We study data, function, and weight anchors, instantiated by generative replay \citep{shin2017continual}, self-distillation \citep{li2017learning}, and importance-based regularization \citep{kirkpatrick2017overcoming,zenke2017continual}, respectively. Low-rank allocation rules determine where successive LoRA updates are retained across tasks \citep{hu2022lora}. We study shared LoRA, which reuses the same adapter across tasks, and merged LoRA, which folds each update into the model before initializing a new adapter.

Testing this hypothesis requires datasets that isolate long-horizon memorization and a method to compare many compositions. Existing public benchmarks for continual language learning primarily measure transfer or performance across heterogeneous downstream tasks or changing corpora \citep{zhang2023citb,wang2023trace,jang2022temporalwiki}. Sequential model-editing benchmarks instead study targeted edits rather than task-wise SFT \citep{hartvigsen2023aging,li2024can}. To address the lack of an appropriate evaluation pipeline, we construct three datasets, each containing 100 tasks, spanning arbitrary symbol associations, LLM-generated fictional facts, and natural questions filtered from public QA datasets. We introduce task-level successive halving to obtain preliminary evidence across many candidate compositions, then use a factorial experiment to measure individual and interaction effects.

Our experiments support the composition hypothesis. Figure~\ref{fig:survival} shows that combining multiple anchors with merged LoRA extends memory lifetime beyond individual mechanisms across all three datasets. After 100 tasks, naive sequential fine-tuning achieves 1.2\% average final retention, measured as accuracy over all learned tasks after the last update, while the best individual mechanism reaches 8.1\%. Our best method combines all three anchors with merged LoRA, which is the only composition in the main factorial that ranks consistently among the top 3 methods in all datasets and achieves 34.9\% average final retention, a 28-fold improvement over naive sequential fine-tuning. The factorial analysis identifies the data anchor and merged LoRA as the largest sources of improvement and finds a super-additive interaction between them on all three datasets.

Our contributions are fourfold. First, we formulate long-horizon continual memorization as a distinct continual learning problem for language models. Second, we organize the design space of mechanism compositions around anchors and low-rank allocation rules. Third, we introduce three 100-task memorization datasets, task-level successive halving, and a factorial evaluation of mechanism combinations. Finally, we show that composing all three anchors with merged LoRA substantially improves retention and ranks among the top 3 methods in all datasets.

\section{Related Work}
\label{sec:related-work}

Catastrophic interference, often called catastrophic forgetting, was first documented in connectionist neural networks and remains a central problem in continual learning \citep{mccloskey1989catastrophic,french1999catastrophic,kirkpatrick2017overcoming}. Contemporary formulations distinguish among task-, domain-, and class-incremental learning according to whether task identity is available at inference time and how the prediction space changes across tasks \citep{van2019three}. Existing methods broadly rely on regularization, replay, or parameter isolation \citep{de2021continual,wang2024comprehensive}. Weight regularization limits changes to parameters that earlier tasks rely on \citep{kirkpatrick2017overcoming,schwarz2018progress,zenke2017continual,aljundi2018memory}, while function regularization preserves earlier model outputs or representations \citep{li2017learning}. Replay uses stored examples \citep{rebuffi2017icarl,rolnick2019experience} or generated samples \citep{shin2017continual}, whereas parameter isolation assigns different model capacity to different tasks \citep{rusu2016progressive,mallya2017packnet}. Some methods combine these signals. Dark Experience Replay jointly uses stored examples and their earlier logits, while Momentum Knowledge Distillation adds a teacher constraint to online continual-learning methods \citep{buzzega2020dark,michel2023rethinking}.

Continual learning methods were largely developed on sequential image classification benchmarks based on MNIST, CIFAR, and ImageNet \citep{lecun1998mnist,krizhevsky2009learning,deng2009imagenet}. Later work extends regularization, replay, and benchmarking to sequential language modeling, instruction tuning, and multitask learning \citep{sun2019lamol,scialom2022fine,zhang2023citb,wang2023trace,xiang2023language}. Continual pretraining instead adapts language models as new corpora, domains, or time periods become available \citep{ke2023continual,ibrahim2024simple,jin2022lifelong,qin2022elle,jang2022temporalwiki}. Low-Rank Adaptation (LoRA) freezes the pretrained model weights and injects trainable rank decomposition matrices into the model to increase training efficiency while better preserving prior knowledge \citep{hu2022lora}. ReLoRA provides a related mechanism for accumulating low-rank updates: during pretraining, it repeatedly merges them into the model and reinitializes the low-rank matrices \citep{lialin2024relora}. Other LoRA-based methods aim to reduce interference across tasks. O-LoRA assigns each task a new update subspace and discourages overlap with earlier subspaces during training \citep{wang2023orthogonal}. OSRM instead uses task features to choose update subspaces that reduce interference when merging independently trained task models \citep{zhang2025unraveling}. Sequential model editing addresses a related problem by asking whether a language model can retain many targeted corrections. Existing studies develop explicit storage for successive edits and show that repeated editing can weaken earlier edits and damage other model capabilities \citep{hartvigsen2023aging,li2024can,gupta2024model}. These memory retention challenges also arise for agents interacting with an environment. AgentOdyssey evaluates test-time continual learning agents in procedurally generated text games, with diagnostic tests of world knowledge acquisition and episodic memory~\citep{zhang2026agentodyssey}.

We instead study long-horizon memorization by measuring the recall of associations across one hundred sequential query-answer tasks in the domain-incremental setting, without task identifiers at inference. Rather than introducing another standalone mechanism, we compose generative replay at the data level, self-distillation at the function level, and importance-based regularization at the parameter level. We combine these anchors with merged LoRA and show that the resulting method substantially outperforms each single mechanism.

\section{Composing Continual Learning Mechanisms}
\label{sec:method}

\subsection{Long-Horizon Memorization via Continual SFT}

We consider an autoregressive language model $p_\Theta$, parametrized by $\Theta_0$, and a sequence of $T$ supervised tasks that arrive one at a time. Task $t$ contains training data $\mathcal{D}_t$, where each example consists of a query $x$ and its target answer $y$. When task $t$ arrives, the model has access to $\mathcal{D}_t$ and inherits the state $S_{t-1}$, including the model parameters $\Theta_{t-1}$. It may not retain or revisit raw training examples from earlier tasks. Let $\Theta_t$ denote the model after learning task $t$. The main challenge of continual supervised fine-tuning (SFT) is to reduce catastrophic forgetting while maintaining the plasticity needed to learn new tasks. The current-task SFT loss is represented by:
\begin{equation}
    \mathcal{L}_{\mathrm{SFT}}^{t}(\Theta)
    =
    -\mathbb{E}_{(x,y)\sim\mathcal{D}_t}
    \left[\log p_\Theta(x,y)\right].
    \label{eq:sft}
\end{equation}
Here $p_\Theta(x,y)$ is the likelihood of the training sequence. Appendix~\ref{app:implementation} specifies the token-level loss. We do not mask the query tokens because, in practice, it is hard to separate the query from the answer in applications such as test-time training.

Data, function, and weight anchors defined in Section~\ref{sec:anchors} add three retention terms to this objective:
\begin{equation}
    \Theta_t
    =
    \operatorname*{arg\,min}_\Theta
    \mathcal{L}_{\mathrm{SFT}}^{t}(\Theta)
    +\mathcal{R}_D^t(\Theta)
    +\mathcal{R}_F^t(\Theta)
    +\mathcal{R}_W^t(\Theta).
    \label{eq:master}
\end{equation}
Here $\mathcal{R}_D^t$, $\mathcal{R}_F^t$, and $\mathcal{R}_W^t$ correspond to the data, function, and weight anchors. Appendix~\ref{app:composition} presents the full objective used to combine the anchors. Low-rank allocation determines which low-rank parameters are updated and what is retained for future tasks. Section~\ref{sec:allocation} describes the low-rank allocation rules in detail.

\subsection{Three anchors}
\label{sec:anchors}

\paragraph{Data anchor.} A data anchor trains the model on replayed sequences that represent earlier tasks. Let $Q_{t-1}$ be a distribution over these sequences, and let $\ell_D(\Theta,z)$ denote the loss applied to $z\sim Q_{t-1}$. Its general form is
\begin{equation}
    \mathcal{R}_D^t(\Theta)
    =
    \mathbb{E}_{z\sim Q_{t-1}}
    \left[\ell_D(\Theta,z)\right].
    \label{eq:data-anchor}
\end{equation}

Vanilla data replay and generative replay construct $Q_{t-1}$ in different ways, while hard and soft replay use different choices of $\ell_D$. Deep Generative Replay uses separate generator and solver networks \citep{shin2017continual}. LAMOL uses a single language model but treats sampled pseudo-sequences as hard training targets \citep{sun2019lamol}.

Our data anchor uses a frozen copy of the previous model to generate complete pseudo-sequences from a single task-agnostic replay token. Before each task after the first, we generate 300 sequences and discard empty outputs. During training, each current-task minibatch is paired with one replay minibatch. The replay weight balances the current-task and replay losses, while the generation temperature controls the randomness of replay sampling. We vary both in the task-level successive-halving study described in Section~\ref{sec:search}. Appendix~\ref{app:hyperparameters} reports the selected values. The frozen model also provides soft next-token targets for the replay sequences, which are used only while learning the current task. Appendix~\ref{app:data-anchor} provides the full generation and replay objectives.

\paragraph{Function anchor.} A function anchor constrains constrains the update of the current model on current-task inputs by comparing its predictions with a reference distribution. Let $\mu_t$ denote the distribution of current-task inputs, let $q_{t-1}(\cdot\mid x)$ denote the reference distribution for input $x$, and let $d$ measure their difference. Its general form is
\begin{equation}
    \mathcal{R}_F^t(\Theta)
    =
    \mathbb{E}_{x\sim\mu_t}
    \left[
      d\!\left(
        q_{t-1}(\cdot\mid x),
        p_\Theta(\cdot\mid x)
      \right)
    \right].
    \label{eq:function-anchor}
\end{equation}
Our function anchor uses the previous model to define the reference distribution, following Learning without Forgetting \citep{li2017learning}. Although the data anchor also uses soft targets, it applies them to generated replay sequences, whereas the function anchor applies self-distillation only to current-task data. Appendix~\ref{app:function-anchor} gives the full self-distillation objective.

\paragraph{Weight anchor.} A weight anchor constrains updates to model parameters according to their estimated importance for previously learned behavior. Let $\vartheta$ denote the parameters of $\Theta$ tracked by the weight anchor, let $\vartheta_{t-1}^{\star}$ be its value before task $t$, and let $H_{t-1}$ encode the accumulated importance. Its general form is
\begin{equation}
    \mathcal{R}_W^t(\Theta)
    =
    \frac{1}{2}
    \bigl(\vartheta-\vartheta_{t-1}^{\star}\bigr)^\top
    H_{t-1}
    \bigl(\vartheta-\vartheta_{t-1}^{\star}\bigr),
    \qquad H_{t-1}\succeq0 .
    \label{eq:weight-anchor}
\end{equation}
EWC applies this quadratic separately for each previous task, using diagonal Fisher information as the importance weights \citep{kirkpatrick2017overcoming}. Online EWC replaces the growing set of task-specific penalties with one penalty centered at the latest parameters using a running Fisher \citep{schwarz2018progress}. SI uses the same diagonal quadratic but estimates importance from contributions accumulated along the optimization path \citep{zenke2017continual}. Appendices~\ref{app:si} and~\ref{app:online-ewc} detail the estimators and hyperparameter settings.

\subsection{Low-Rank Allocation}
\label{sec:allocation}

Anchors constrain the current update. A low-rank allocation rule determines which low-rank parameters are used for each task and how the learned update is retained for later tasks. For a pretrained matrix $W_0$, LoRA can be specified as $\rho BA$, where $A\in\mathbb{R}^{r\times d_{\mathrm{in}}}$, $B\in\mathbb{R}^{d_{\mathrm{out}}\times r}$, and $\rho=\alpha_{\mathrm{LoRA}}/r$ \citep{hu2022lora}. Let $A_t$ and $B_t$ denote the LoRA matrices optimized during task $t$, and let a superscript $\star$ denote their values after training on task t. We mainly consider two ways to carry these matrices across tasks:
\begin{equation}
    W_t
    =
    \begin{cases}
        W_0+\rho B_tA_t,
        & \text{shared LoRA},\\
        W_{t-1}+\rho B_tA_t,
        & \text{merged LoRA}.
    \end{cases}
    \label{eq:allocation}
\end{equation}

Shared LoRA continues optimizing the same $A$ and $B$ matrices across all tasks, so $B_tA_t$ represents the single complete LoRA adapter after learning tasks $1$ through $t$. Merged LoRA instead assigns each task a new pair of LoRA matrices. After task $t$, it folds $\rho B_t^\star A_t^\star$ to the dense matrix $W_{t-1}$ and initializes a new pair of LoRA matrices for the next task. This rule adapts ReLoRA's merge and reinitialization pattern to continual learning by merging the LoRA update into the dense weights after each task and initializing new LoRA matrices and a new optimizer for the next task \citep{lialin2024relora}. Both methods retain one dense model and one pair of LoRA matrices per adapted weight matrix, so their retained state size remains constant as the number of tasks increases.

The main experiments compare different anchor combinations using shared LoRA and merged LoRA. Appendix~\ref{app:res-allocation} reports separate experiments with O-LoRA and sequential OSRM, whose state grows with the number of tasks. Appendices~\ref{app:shared-allocation}--\ref{app:osrm} detail the update rules and state complexity of all four methods.

\section{Evaluation Setup}
\label{sec:evaluation}

\subsection{Protocol and metrics}

We follow the domain-incremental setting of \citet{van2019three} that
the model receives one task at a time and is not given the task identity during inference. Each task contains query-answer pairs. Because we study memorization rather than generalization, each task is evaluated on the same examples used for training.
Each dataset is an ordered stream of $T=100$ tasks. After learning task $i$, we evaluate the model on every task $j\leq i$. Let $x_{j,n}$ and $y_{j,n}$ denote the query and answer for example $n$ in task $j$, and let $N_j$ be the number of examples in that task. We write $\hat y_i(x)$ for the answer produced by the model after learning task $i$. The temporal accuracy matrix is
\begin{equation}
    M_{i,j}
    =
    \frac{1}{N_j}\sum_{n=1}^{N_j}
    \mathbf{1}\!\left[
      \hat y_i(x_{j,n})=y_{j,n}
    \right],
    \qquad 1\leq j\leq i\leq T.
    \label{eq:matrix}
\end{equation}
From this matrix, we report final retention, immediate acquisition, and forgetting:
\begin{equation}
\begin{aligned}
    \finalacc &= \frac{1}{T}\sum_{j=1}^{T}M_{T,j}, &
    \diagacc &= \frac{1}{T}\sum_{j=1}^{T}M_{j,j}, &
    \forget &= \frac{1}{T-1}\sum_{j=1}^{T-1}
    \left(\max_{j\leq i\leq T}M_{i,j}-M_{T,j}\right).
\end{aligned}
\label{eq:metrics}
\end{equation}
$\finalacc$ is the mean accuracy over all tasks after task $T$. $\diagacc$ is the mean accuracy on each task immediately after it is learned. $\forget$ is the average drop from each task's best observed accuracy to its final accuracy. It excludes the last task because no later update can cause it to be forgotten.

\subsection{Three memorization datasets}

Our three datasets increase in semantic realism. \symbolqa{} contains 10,000 random key-value associations. \llmqa{} contains 10,000 query-answer pairs generated by an LLM across 100 fictional topics. For these synthetically generated datasets, we ensure that each query maps to exactly one target answer across all tasks. \realqa{} contains 5,000 natural query-answer pairs from ten public QA datasets, filtered to exclude items the model answers correctly in any of five sampled completions. Each dataset contains 100 tasks, with 100 examples per task for \symbolqa{} and \llmqa{}, and 50 for \realqa{}. Appendix~\ref{app:data} gives the source list and full construction of each dataset. Appendices~\ref{app:implementation}--\ref{app:hyperparameters} provide the method definitions and experimental settings.

\subsection{Searching the Combinatorial Design Space}
\label{sec:search}

Crossing the three anchor categories in Section~\ref{sec:anchors} with the low-rank allocation rules in Section~\ref{sec:allocation} yields many possible continual learning methods. We use this design space to seek preliminary evidence for our hypothesis. Training every method on all 100 tasks in each dataset is expensive, but rankings after only a few tasks may not reliably identify which methods will perform best at longer task horizons. We therefore introduce task-level successive halving (TSH) over progressively longer task horizons.

\paragraph{Task-Level Successive Halving.}
Unlike prior applications of successive halving that allocate increasing numbers of training iterations to promising hyperparameters \citep{jamieson2016non}, our TSH increases the number of sequential tasks and rank configurations by retention at each task horizon. Here, a task horizon $r$ refers to the number of tasks that a configuration of method has learned.

Denote the full search space of method compositions by $\mathcal{A}_1$, and let $\mathcal{S}$ be the set of training seeds. For configuration $a\in\mathcal{A}_1$ and seed $s\in\mathcal{S}$, let $M_{i,j}^{a,s}$ be the resulting temporal accuracy matrix, where $M_{i,j}^{a,s}$ is the accuracy on task $j$ after learning tasks $1$ through $i$. Every seed in $\mathcal{S}$ uses the same task order, fixed by the separate task-order seed, where we call the development task order. Averaging over $\mathcal{S}$ therefore captures training stochasticity but not sensitivity to task order. After $r$ tasks, we score each configuration by its mean final retention across these seeds:
\begin{equation}
    F_r(a)=\frac{1}{|\mathcal{S}|}\sum_{s\in\mathcal{S}}\frac{1}{r}\sum_{j=1}^{r}M_{r,j}^{a,s}.
    \label{eq:search-score}
\end{equation}

Let $\varnothing$ indicate that an anchor is absent. The initial candidate set is
\begin{equation}
\begin{aligned}
\mathcal{A}_1={}&\{\varnothing,\text{online EWC},\text{SI}\}
\times\{\varnothing,\text{SD}_1,\text{SD}_2\}\\
&\times\{\varnothing,\text{Replay}_1,\ldots,\text{Replay}_4\}
\times\{\text{shared LoRA},\text{merged LoRA}\},\\
n_1={}&|\mathcal{A}_1|=3\times3\times5\times2=90.
\end{aligned}
\label{eq:candidate-set}
\end{equation}
$\text{SD}_1$ and $\text{SD}_2$ use self-distillation loss weights 1 and 3. $\text{Replay}_1,\ldots,\text{Replay}_4$ enumerate the Cartesian product of replay loss weights $\{0.5,0.75\}$ and generation temperatures $\{1.0,1.5\}$. All other optimization settings remain fixed. Appendix~\ref{app:hyperparameters} gives the complete settings.

Starting with all 90 configurations, we retain the top 45 after 10 tasks, the top 23 after 20 tasks, and the top 10 after 50 tasks. These final ten configurations continue through all 100 tasks. TSH therefore uses early retention to decide which configurations receive further training. Although this procedure does not guarantee that it retains the best configuration, the 10-task search rankings show strong agreement with the 100-task final-evaluation rankings for the method compositions evaluated in both phases, despite the different task orders. Algorithm~\ref{alg:search} presents the detailed procedure. Appendix~\ref{app:search-cost} provides the resource accounting, and Appendix~\ref{app:search-order} reports the ranking comparisons.

\begin{figure}[!t]
\centering
\includegraphics[width=\linewidth]{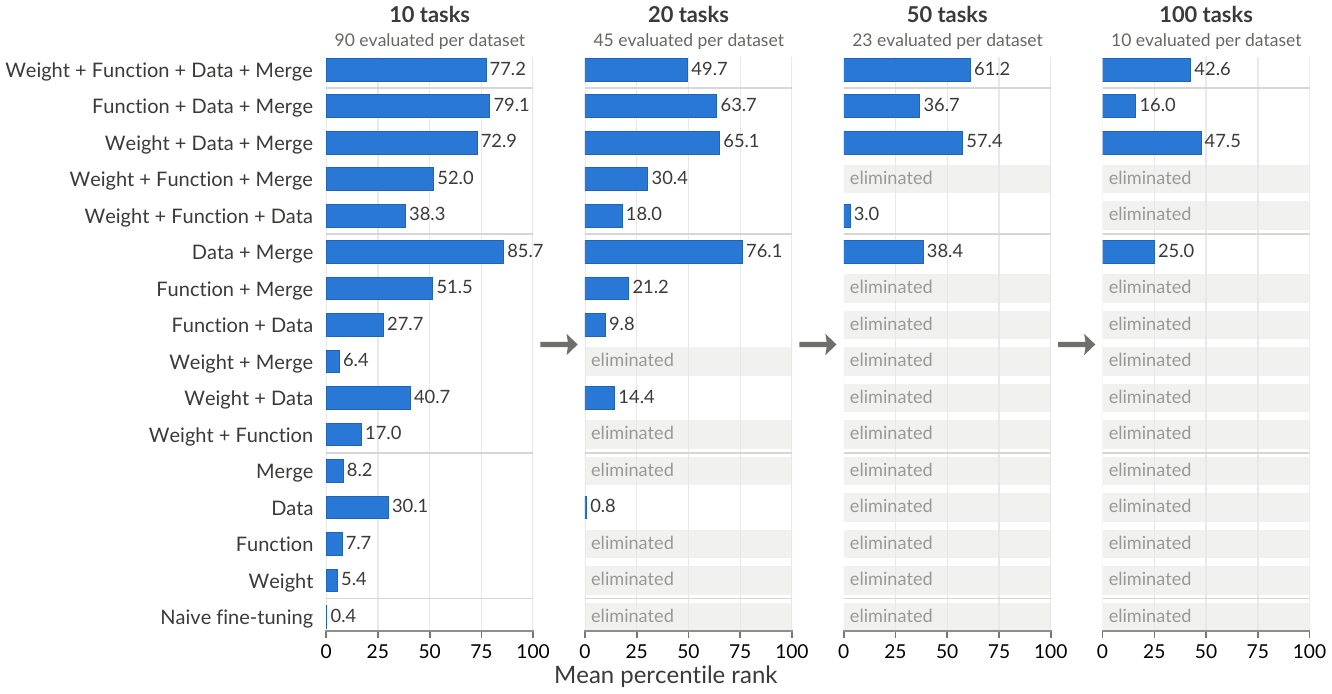}
\caption{Task-level successive halving. Within each dataset and horizon, we rank surviving candidates by mean retention over three seeds. A method's percentile rank is the percentage of other methods ranked below it. Bars average these values over method variants and then over the three datasets. Because the candidate pool changes, bar heights do not measure changes in retention across horizons. Gray entries indicate that all variants have been eliminated in all three datasets. The leading methods at 100 tasks combine multiple anchors with merged LoRA.}
\label{fig:sho-results}
\end{figure}

\section{Experiments and Results}
\label{sec:results}

Figure~\ref{fig:sho-results} shows that no standalone mechanism reaches the 50-task stage of TSH, while every method reaching 100 tasks combines a data anchor with merged LoRA. The winner on each dataset also includes a weight anchor, and the \symbolqa{} winner additionally uses a function anchor. These results provide preliminary support for combining multiple anchors with merged LoRA. We therefore evaluate all combinations of the three anchors and merged LoRA in a $2^4$ factorial to measure their individual and interaction effects. We use the hyperparameters selected by TSH and the default 100-task order, which differs from the search order. Our primary focus is methods whose retained state remains constant as the number of tasks increases. To test whether state that grows with the number of tasks improves retention, we evaluate O-LoRA and sequential OSRM both individually and as replacements for merged LoRA in each dataset's TSH winner. The full evaluation contains 21 methods per dataset, each with three seeds.

\begin{figure}[!t]
\centering
\includegraphics[width=\linewidth]{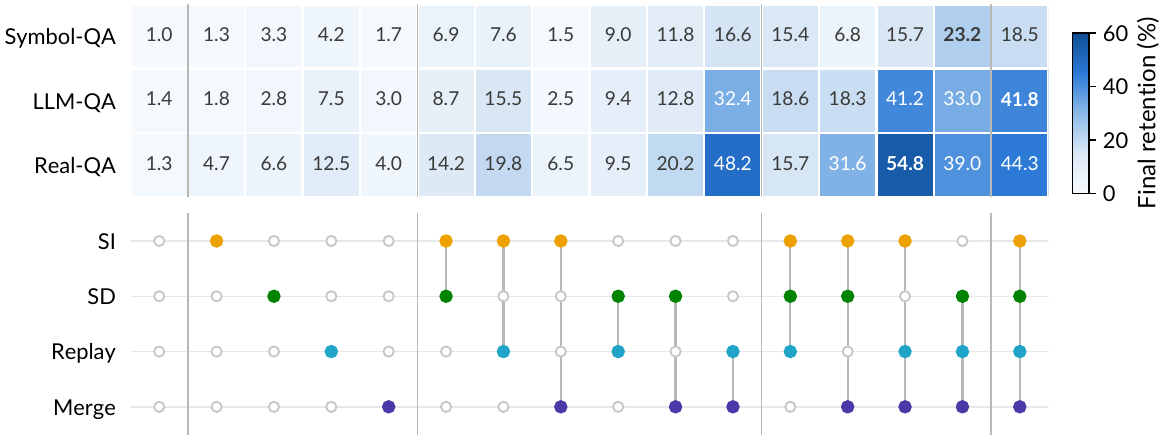}
\caption{Final retention (\%) after 100 tasks for all 16 combinations, averaged over three seeds. Filled markers identify active mechanisms, and an inactive Merge marker indicates shared LoRA. The leftmost column is shared LoRA without anchors. Bold marks the best result per dataset. Appendix~\ref{app:res-all-configs} reports standard deviations over seeds. The best compositions substantially outperform standalone mechanisms.}
\label{fig:composition-results}
\end{figure}

\subsection{Compositions outperform standalone mechanisms}
\label{sec:compositions}

Across the full factorial results in Figure~\ref{fig:composition-results}, compositions are substantially more effective than any standalone mechanism after 100 tasks. The strongest standalone mechanism retains only 4.2\% on \symbolqa{}, 7.5\% on \llmqa{}, and 12.5\% on \realqa{}. By contrast, the highest mean retention among the compositions reaches 23.2\%, 41.8\%, and 54.8\% on the three datasets. The strongest composition also varies by dataset. \symbolqa{} favors SD with replay and merged LoRA, \llmqa{} favors the full stack of SI, SD, replay, and merged LoRA, and \realqa{} favors SI with replay and merged LoRA. Despite this variation, our best method, which combines all three anchors with merged LoRA, is the only composition that ranks among the top 3 methods in all datasets. It achieves 34.9\% average final retention across the three datasets. Appendix Table~\ref{tab:app-factorial-ranks} reports the complete cross-dataset ranking. Appendix~\ref{app:res-all-configs} shows that these methods achieve nearly perfect immediate acquisition, so the differences in final retention primarily reflect forgetting rather than difficulty learning the current task. Appendix~\ref{app:res-buildup} also shows how retention changes along a greedy path through the factorial.

Replacing merged LoRA with O-LoRA changes final retention only slightly, with small gains on the two natural-language datasets and a decrease on \symbolqa{}. Sequential OSRM lowers retention on all three datasets. Thus, low-rank allocation rules with task-growing state do not consistently improve retention. Appendix~\ref{app:res-allocation} reports this comparison in detail. We evaluate general capability on held-out mathematical reasoning and knowledge benchmarks: GSM8K~\citep{cobbe2021training}, MATH~\citep{hendrycks2021measuring}, MGSM~\citep{shi2022language}, and MMLU-Redux~\citep{gema2025we}. Improved memorization does not necessarily preserve general capability. All methods exhibit catastrophic forgetting on these benchmarks. However, O-LoRA preserves substantially more general capability than merged LoRA after training on \llmqa{} and \realqa{}, despite only small gains in final retention. Appendix~\ref{app:res-capability} reports the full comparison.

\subsection{Composition extends memory half-life}
\label{sec:res-time}

Final retention summarizes only performance after the last task. To examine how memories decay during training, we group the temporal accuracy matrix by memory age. For age $a$, $R(a)$ averages all entries $M_{i,j}$ with $i-j=a$, and therefore measures accuracy after $a$ later tasks have been learned. We define memory half-life as the first age at which $R(a)$ falls below half of $R(0)$. Figure~\ref{fig:survival} compares these trajectories for naive fine-tuning, the strongest standalone mechanism, our best method, and the strongest composition for each dataset.

Naive fine-tuning has a memory half-life of only one task on \symbolqa{} and \llmqa{}, and two tasks on \realqa{}. The strongest standalone mechanism extends these half-lives to 4, 6, and 11 tasks. The strongest composition extends them further to 19, 32, and 44 tasks. Our best method reaches 19, 32, and 32 tasks, matching the strongest composition on the first two datasets and remaining substantially stronger than any standalone mechanism on \realqa{}. Even so, all curves continue to decline with memory age. Composition therefore delays forgetting by changing its timescale, but it does not prevent the eventual loss of older memories.

\subsection{Replay and merged LoRA provide the largest gains}
\label{sec:res-factorial}

Table~\ref{tab:anova} reports the main effects of the four mechanisms and all interactions among them on final retention. Appendix~\ref{app:res-anova} provides the full test statistics.

\begin{table}[!t]
\centering
\setlength{\tabcolsep}{4pt}
\renewcommand{\arraystretch}{1.15}
\begin{tabular*}{\linewidth}{@{\extracolsep{\fill}}lrrrrrrrrrr@{}}
\toprule
& \multicolumn{4}{c}{Main effects}
& \multicolumn{6}{c}{Two-way}
\\
\cmidrule(lr){2-5}\cmidrule(lr){6-11}
Dataset
& SI & SD & R & M
& $\mathrm{SI}{\times}\mathrm{SD}$
& $\mathrm{SI}{\times}\mathrm{R}$
& $\mathrm{SI}{\times}\mathrm{M}$
& $\mathrm{SD}{\times}\mathrm{R}$
& $\mathrm{SD}{\times}\mathrm{M}$
& $\mathrm{R}{\times}\mathrm{M}$
\\
\midrule
\symbolqa{}
& $+0.3$ & $\mathbf{+5.7}$ & $\mathbf{+9.5}$ & $\mathbf{+5.9}$
& $-0.3$ & $+0.7$ & $\mathbf{-3.0}$ & $-0.2$ & $+0.6$ & $\mathbf{+3.6}$
\\
\llmqa{}
& $\mathbf{+5.8}$ & $\mathbf{+5.0}$ & $\mathbf{+18.5}$ & $\mathbf{+14.9}$
& $+1.6$ & $\mathbf{+2.9}$ & $-0.1$ & $\mathbf{-3.5}$ & $+1.7$ & $\mathbf{+9.4}$
\\
\realqa{}
& $\mathbf{+6.3}$ & $\mathbf{+3.7}$ & $\mathbf{+19.3}$ & $\mathbf{+20.5}$
& $+1.3$ & $+0.1$ & $+0.2$ & $\mathbf{-10.3}$ & $+1.7$ & $\mathbf{+11.7}$
\\
\addlinespace[2pt]
\midrule
\end{tabular*}
\vspace{0.45em}
\begin{tabular*}{\linewidth}{@{\extracolsep{\fill}}lrrrrr@{}}
& \multicolumn{4}{c}{Three-way}
& \multicolumn{1}{c}{Four-way} \\
\cmidrule(lr){2-5}\cmidrule(lr){6-6}
Dataset
& $\mathrm{SI}{\times}\mathrm{SD}{\times}\mathrm{R}$
& $\mathrm{SI}{\times}\mathrm{SD}{\times}\mathrm{M}$
& $\mathrm{SI}{\times}\mathrm{R}{\times}\mathrm{M}$
& $\mathrm{SD}{\times}\mathrm{R}{\times}\mathrm{M}$
& $\mathrm{SI}{\times}\mathrm{SD}{\times}\mathrm{R}{\times}\mathrm{M}$ \\
\midrule
\symbolqa{} & $+0.1$ & $\mathbf{-1.8}$ & $-0.8$ & $-1.3$ & $+0.2$ \\
\llmqa{} & $-1.3$ & $-0.1$ & $+0.2$ & $\mathbf{-2.6}$ & $-0.2$ \\
\realqa{} & $-1.9$ & $+0.5$ & $-0.6$ & $\mathbf{-4.9}$ & $-0.6$ \\
\addlinespace[2pt]
\bottomrule
\end{tabular*}
\caption{Main and interaction effects from the $2^4$ factorial analysis of final retention, in percentage points. R is replay and M is merged LoRA. Bold marks statistically significant entries ($p<0.05$). Replay and merged LoRA have the largest main effects and a positive interaction in all three datasets.}
\label{tab:anova}
\end{table}

Replay and merged LoRA have the two largest main effects on every dataset. These main effects measure the average change in final retention from adding a mechanism across all configurations of the other three mechanisms. Replay raises retention by 9.5 to 19.3 percentage points, while merged LoRA raises it by 5.9 to 20.5 points. Their combination is also strongly super-additive. The interaction between replay and merged LoRA is positive and significant on all three datasets. Figure~\ref{fig:composition-results} shows this synergy directly in the configurations without SI or SD. Across \symbolqa{}, \llmqa{}, and \realqa{}, the standalone gains from replay and merged LoRA sum to only 3.9, 7.7, and 13.9 points, respectively. Combining them instead improves retention over naive fine-tuning by 15.6, 31.0, and 46.9 points. The gains from the pair therefore far exceed the sum of their standalone contributions.

The contributions of SI and SD depend more strongly on the dataset and the other composition. SD has a positive main effect on every dataset, but its negative interaction with replay on \llmqa{} and \realqa{} shows that its average benefit is smaller when replay is already present. SI has positive main effects on the two natural-language datasets but no detectable main effect on \symbolqa{}. Its interaction with merged LoRA is negative and significant only on \symbolqa{}. As discussed in Appendix~\ref{app:coordinate-compat}, merged LoRA replaces the LoRA factors after each task, while SI carries forward importance values and references tied to the previous factors. These values are therefore applied to newly initialized coordinates whose functional roles have changed. The negative interaction on \symbolqa{} is consistent with this mismatch. Its absence on the natural-language datasets shows that the mismatch creates a structural risk rather than a universal empirical penalty. One possible explanation is that the arbitrary mappings in \symbolqa{} provide less reusable structure across tasks, making misplaced constraints on the fresh workspace more costly.

Overall, replay and merged LoRA form the common core of strong compositions. SI and SD can provide additional gains, but their value depends on both the dataset and the other active mechanisms.

\section{Conclusion}
\label{sec:conclusion}

We formalize long-horizon continual memorization and organize its design space around data, function, and weight anchors and low-rank allocation rules. We introduce three 100-task query-answer datasets of increasing naturalness, task-level successive halving to obtain early evidence about mechanism composition, and a factorial design to measure individual and interaction effects. No individual mechanism retains knowledge well after 100 tasks. Our best method combines all three anchors with merged LoRA and is the only factorial composition that ranks among the top 3 methods in all datasets. It raises average final retention from 1.2\% under naive fine-tuning to 34.9\%, a 28-fold improvement. The factorial analysis identifies the data anchor and merged LoRA as the largest sources of improvement and shows a super-additive interaction between them on every dataset. These results show that effective long-horizon continual memorization depends on finding the right combination of complementary mechanisms.

\paragraph{Memorization rather than generalization.} Our evaluation tests recall using the queries seen during training. A model may retain the corresponding associations yet answer a paraphrased query incorrectly. Our results therefore do not establish generalization to new query formulations.

\paragraph{General capability preservation.} Stronger memorization does not ensure preservation of general capability. The evaluated methods still lose substantial accuracy on general capability benchmarks after 100 tasks (Appendix~\ref{app:res-capability}). Preserving these abilities while learning new associations remains an open challenge.

\bibliography{references}

\clearpage
\appendix
\section{Extended Related Work}
\label{app:related-work}

\paragraph{Continual-learning settings.} Continual learning trains one model on a sequence of tasks while seeking to preserve performance on earlier data. Standard formulations distinguish task-, class-, and domain-incremental learning according to how the output space changes and whether inference provides a task identifier \citep{van2019three}. Our setting uses the same question--answer interface and vocabulary throughout training and provides no task identifier at evaluation. It therefore tests whether one model can retain all associations without routing each input to a task-specific predictor.

\paragraph{Mechanisms for preserving earlier knowledge.} Regularization methods constrain learning through information retained from previous tasks. Weight-based methods assign importance to individual parameters and penalize changes to important values, as in EWC, online EWC, SI, and MAS \citep{kirkpatrick2017overcoming,schwarz2018progress,zenke2017continual,aljundi2018memory}. Function-based methods instead constrain model behavior. Learning without Forgetting, for example, asks the current model to match a previous model's outputs on available inputs \citep{li2017learning}. Replay methods train on earlier examples stored in memory \citep{rebuffi2017icarl,rolnick2019experience} or on samples generated to approximate earlier data \citep{shin2017continual}. Parameter-isolation methods reduce interference by assigning different parameters to different tasks or by restricting which parameters each task may change \citep{rusu2016progressive,mallya2017packnet}. Our data, function, and weight anchors instantiate the first three preservation signals, while low-rank allocation rules determine whether successive task updates reuse or separate low-rank capacity.

\paragraph{Combining continual-learning mechanisms.} Prior work combines preservation signals, but it does not systematically study the broader space of mechanism compositions considered here. Dark Experience Replay stores examples together with the logits produced when those examples entered memory, then uses both rehearsal and output matching during later learning \citep{buzzega2020dark}. Momentum Knowledge Distillation adds a slowly updated teacher to online continual learning methods and studies how distillation complements replay \citep{michel2023rethinking}. Our study also examines interactions among mechanisms, but it uses generated replay rather than stored examples and evaluates data, function, and weight anchors both individually and in controlled compositions with different low-rank allocation rules.

\paragraph{Continual learning for language models.} Prior language work studies sequential language modeling, task learning, and instruction tuning. LAMOL learns to answer current-task examples and generate pseudo-examples for earlier tasks \citep{sun2019lamol}. Later work examines continual fine-tuning, instruction-tuning benchmarks, and forgetting across diverse language tasks \citep{scialom2022fine,zhang2023citb,wang2023trace,xiang2023language}. Continual pretraining instead updates a language model as new corpora, domains, or time periods arrive \citep{jin2022lifelong,qin2022elle,jang2022temporalwiki,ke2023continual,ibrahim2024simple}. These studies primarily measure language modeling, transfer, or downstream task performance. We isolate associative retention by evaluating the same question--answer items before and after many later tasks.

REPINA mitigates representation collapse during fine-tuning by matching fine-tuned representations to pretrained representations, either directly or through a learned projection~\citep{razdaibiedina2023representation}. Our function anchor instead matches the previous model's output distributions on current-task data.

\paragraph{Sequential model editing.} Sequential model editing also asks a model to retain many updates. GRACE leaves the original model weights unchanged and stores edits in a discrete key--value codebook that activates for inputs near a stored key \citep{hartvigsen2023aging}. Other studies show that applying model editors repeatedly can weaken earlier edits, reduce the model's ability to learn new edits, and harm downstream performance \citep{li2024can,gupta2024model}. This literature emphasizes edit success, generalization to related inputs, and limited changes to unrelated behavior. Our setting instead presents sets of associations as ordinary training data and measures recall across one hundred task boundaries. The two settings share the problem of retaining many factual updates, but they differ in how they introduce and evaluate those updates.

\paragraph{Low-rank allocation.} LoRA represents an update to a frozen weight matrix with two low-rank matrices that can be merged into the dense weight for evaluation \citep{hu2022lora}. ReLoRA repeatedly merges and reinitializes low-rank matrices during pretraining, allowing multiple low-rank updates to produce a higher-rank cumulative change \citep{lialin2024relora}. Continual learning methods use this structure in several ways. CoLoR trains a separate LoRA expert for each task and infers which expert to use at evaluation \citep{wistuba2023continual}. O-LoRA retains the matrices learned for earlier tasks and penalizes overlap between the current and earlier $A$ matrices \citep{wang2023orthogonal}. InfLoRA constructs task-specific update subspaces to reduce interference with earlier tasks \citep{liang2024inflora}. OSRM uses task features to initialize LoRA subspaces before independently training and later merging task models \citep{zhang2025unraveling}. Our comparison separates the low-rank allocation rule from the preservation objective. Shared LoRA reuses one pair of matrices, merged LoRA commits each task's update to the dense weights before creating a new pair, O-LoRA retains a new pair for each task, and our sequential OSRM adaptation combines merged LoRA with an initialization derived only from completed tasks.

\section{Detailed Method Implementations}
\label{app:methods}

\subsection{Scope, taxonomy, and notation}
\label{app:implementation}

This appendix specifies the methods in our evaluation suite. We distinguish anchors, which preserve evidence about earlier tasks, from low-rank allocation, which specifies the low-rank parameters used for each task and how learned updates are retained. Vanilla sequential SFT is the no-anchor baseline and uses shared LoRA as its low-rank allocation rule. The following subsections describe each anchor and low-rank allocation rule. The data anchor is unconditional generative replay~\ref{app:data-anchor}, the function anchor is self-distillation~\ref{app:function-anchor}, the weight anchors are Synaptic Intelligence (SI)~\ref{app:si}, and the online EWC~\ref{app:online-ewc}. The low-rank allocation rules are shared LoRA~\ref{app:shared-allocation}, merged LoRA~\ref{app:merge-allocation}, O-LoRA~\ref{app:olora}, and our strictly sequential adaptation of OSRM~\ref{app:osrm}. Table~\ref{tab:method-taxonomy} summarizes this categorization.

\begin{table}[H]
    \centering
    \small
    \begin{tabular}{p{0.16\linewidth}p{0.22\linewidth}p{0.31\linewidth}p{0.20\linewidth}}
        \hline
        Category & Method & Persistent learner state & Horizon dependence \\
        \hline
        Vanilla baseline &
        Sequential SFT (no anchor) &
        one shared LoRA workspace
        & Constant \\
        Data anchor & Unconditional generative replay & Model trained from last task to generate replay data unconditionally. No raw examples or replay samples are carried to the next task & Constant \\
        Function anchor &  self-distillation & Model trained from the last task treated as a frozen teacher during the current task for KL estimation & Constant \\
        Weight anchor & Synaptic Intelligence & One reference and one diagonal importance tensor per trainable tensor saved after each task and then discarded before training the next task & Constant \\
        Weight anchor & Online EWC & One reference and one running diagonal Fisher tensor per trainable tensor saved after each task and then discarded before training the next task & Constant \\
        Low-rank allocation & shared LoRA & One continually updated rank-$r$ adapter used by vanilla SFT and composable with any anchor. & Constant \\
        Low-rank allocation & merged LoRA & Trained model from last task only and a fresh rank-$r$ LoRA for the current task & Constant \\
        Low-rank allocation & O-LoRA & Frozen rank-$r$ factors for every completed task and one fresh rank-$r$ LoRA & Linear in tasks \\
        Low-rank allocation & Sequential OSRM & Trained model from last task only, a fresh rank-$r$ LoRA for the current task, and one feature vector per past task and target module & Linear in tasks \\
        \hline
    \end{tabular}
    \caption{Methods used in our project, categorized as anchors or low-rank allocation rules. ``Constant'' refers to a method that depends only on a fixed set of information, while ``Linear in tasks'' means the method needs to store information that grows with the number of tasks during training.}
    \label{tab:method-taxonomy}
\end{table}
For task $t$, let $\Theta$ denote the student model being optimized and let $\bar{\Theta}_{t-1}$ denote the frozen model obtained after task $t-1$. A training example is a token sequence $z=(z_1,\ldots,z_L)$ containing the question, the literal answer delimiter, and the boxed answer. When the replay seed is enabled, it is prepended to the sequence. We train on the entire sequence of question and answer pairs since during the deployment phase, we don't usually have the question prefix that we can mask from the SFT loss computation. We mask the replay seed token. For a current-task minibatch $\mathcal{B}$, let $L_n$ denote the number of tokens for minibatch $n$, our implementation computes
\begin{equation}
    \widehat{\mathcal{L}}_{\mathrm{SFT}}^t(\Theta,\mathcal{B})
    =
    -\frac{1}{\sum_{n\in\mathcal{B}}\sum_{\ell=1}^{L_n}m_{n,\ell}}
    \sum_{n\in\mathcal{B}}\sum_{\ell=1}^{L_n}
    m_{n,\ell}
    \log p_{\Theta}(z_{n,\ell}\mid z_{n,<\ell}),
    \label{eq:app-sft}
\end{equation}
where $m_{n,\ell}=1$ at non-padding positions and $0$ otherwise. Equation~\ref{eq:sft} is the expectation of this minibatch loss under data shuffling, so the current-task objective covers the complete sequence. The data-anchor loss instead uses teacher-generated unconditional generative replay sequences and is defined in Section~\ref{app:data-anchor}.

All low-rank allocation rules operate on the same set $\mathcal{J}$ of adapted linear maps. For $j\in\mathcal{J}$, let $W_j^{eff}$ represent the updated LLM layer $j$ parameters with a LoRA module \citep{hu2022lora}, $\alpha_{LoRA}$ and $r$ represent the LoRA $\alpha$ and LoRA ranks, respectively. We can express it as the following:
\begin{equation}
    W_j^{\mathrm{eff}}
    =
    W_j+\rho B_jA_j,
    \qquad
    A_j\in\mathbb{R}^{r\times d_{\mathrm{in},j}},
    \quad
    B_j\in\mathbb{R}^{d_{\mathrm{out},j}\times r},
    \quad
    \rho=\frac{\alpha_{\mathrm{LoRA}}}{r}.
    \label{eq:app-lora}
\end{equation}
During training, dropout is applied to the input of the low-rank branch and is disabled at evaluation. We use Kaiming initialization for $A_j$ and initialize $B_j=0$, so a fresh adapter initially leaves the model function unchanged. Section~\ref{app:allocation-details} describes how each low-rank allocation rule carries these factors across task boundaries.

\subsection{Full compositional objective}
\label{app:composition}

This section states the loss that the optimizer minimizes at task $t$. A configuration selects one low-rank allocation rule and may add anchors. This rule determines which LoRA parameters remain trainable and how the method carries them across task boundaries. Shared LoRA, merged LoRA, and sequential OSRM do not add terms to the loss. O-LoRA also regularizes its factors during training, so its two regularizers appear in the complete objective below. The following subsections give the mathematical details of every anchor and low-rank allocation rule.

We use binary indicators $a_D^t$, $a_F^t$, $a_{\mathrm{SI}}^t$, and $a_{\mathrm{EWC}}^t$ for the data, function, SI, and online EWC anchors. An anchor indicator equals one only when the configuration uses that anchor and the anchor has state from an earlier task. All four anchor indicators therefore equal zero at $t=1$. The indicator $a_{\mathrm{O}}^t$ equals one when the configuration uses O-LoRA. Each $\lambda$ is a nonnegative coefficient that controls the strength of its term.

The data anchor changes the data used to fit the model. Let $\mathcal{L}_{D}^t$ denote the replay loss on sequences that the model from task $t-1$ generates. When the configuration uses replay and an earlier model exists, we mix this loss with the current task SFT loss instead of treating replay as a separate regularizer:
\begin{equation}
    \mathcal{L}_{\mathrm{fit}}^t
    =
    \begin{cases}
        \mathcal{L}_{\mathrm{SFT}}^t,
        & a_D^t=0,\\[2mm]
        (1-w)\mathcal{L}_{\mathrm{SFT}}^t
        +w\mathcal{L}_{D}^t,
        & a_D^t=1.
    \end{cases}
    \label{eq:app-fit}
\end{equation}
Here $w\in[0,1]$ sets the replay ratio. Equation~\ref{eq:app-fit} reduces to ordinary SFT when the data anchor is off or when $t=1$.

The other terms serve different roles. The function anchor loss $\mathcal{L}_{F}^t$ matches the previous model's distribution over the next token on current task sequences. The SI penalty $\mathcal{R}_{\mathrm{SI}}^t$ limits changes to trainable scalar coordinates that contributed strongly along earlier optimization paths. The online EWC penalty $\mathcal{R}_{\mathrm{EWC}}^t$ limits changes to coordinates with a running diagonal Fisher value. For O-LoRA, $\mathcal{R}_{\perp}^t$ discourages overlap between the new input side factors and the factors from earlier tasks, while $\mathcal{R}_{2}^t$ controls the norm of the current factors. With these terms, the optimizer minimizes
\begin{align}
    \mathcal{L}_{\mathrm{impl}}^t
    ={}&
    \mathcal{L}_{\mathrm{fit}}^t
    +a_F^t\lambda_F\mathcal{L}_{F}^t
    +a_{\mathrm{SI}}^t\lambda_{\mathrm{SI}}\mathcal{R}_{\mathrm{SI}}^t
    +a_{\mathrm{EWC}}^t\lambda_{\mathrm{EWC}}\mathcal{R}_{\mathrm{EWC}}^t
    \nonumber\\
    &+
    a_{\mathrm{O}}^t
    \left(
      \lambda_{\perp}\mathcal{R}_{\perp}^t
      +\lambda_2\mathcal{R}_2^t
    \right).
    \label{eq:app-objective}
\end{align}
When the data anchor is active, we pair each current minibatch with one replay minibatch. We create a shuffled replay iterator at the start of each epoch and create another one whenever it is exhausted, so every current minibatch receives a replay partner. Self-distillation is evaluated only on the current minibatch and constrains the current model to match the previous model on current-task sequences. For all low-rank allocation rules except O-LoRA, SI records the gradient of $\mathcal{L}_{\mathrm{fit}}^t$ before adding the function and weight penalties. With O-LoRA, SI records only the weighted current-task loss. In every case, the optimizer updates the model using the complete objective in Equation~\ref{eq:app-objective}.

\subsection{Data anchor: unconditional generative replay}
\label{app:data-anchor}
The data anchor approximates the distribution of earlier tasks using sequences sampled from the previous model. It stores no raw examples from earlier tasks. At the start of each task, the previous model generates a temporary replay set that is used only while learning the current task. Generated samples may be retained as diagnostic logs, but they are not part of the learner state and are never used to train later tasks.

\subsubsection{Seed token}
\label{app:replay-seed}

Let $\mathcal{D}_t$ denote the training data for task $t$. We add one token, $s=\texttt{<|replay\_token|>}$, to the vocabulary and prepend it to every training sequence. For a query-answer pair $(q,a)\in\mathcal{D}_t$, the formatted sequence is
\begin{equation}
    z(q,a)
    =
    s \;\Vert\; \texttt{Question: }\texttt{q}
    \;\Vert\; \texttt{\textbackslash nAnswer: }
    \texttt{\string\boxed\{a\}},
    \label{eq:app-replay-render}
\end{equation}
where $\Vert$ denotes concatenation. Because the same token prefixes the training sequences from every task, the model learns to generate complete formatted sequences from $p_{\Theta}(\cdot\mid s)$. The data anchor samples this distribution instead of storing training data from earlier tasks.

We initialize the input embedding of $s$ to the mean of the original vocabulary embeddings:
\begin{equation}
    E[s]
    =
    \frac{1}{|V|}\sum_{v\in V}E[v],
    \label{eq:app-replay-init}
\end{equation}
where $V$ is the original vocabulary and $E$ is the input embedding matrix. When the input and output embeddings are not tied, we initialize the corresponding output embedding by the same rule. The replay-token embeddings remain frozen. During evaluation, we prepend $s$ to prompts for every checkpoint trained with the replay token so that training and evaluation use the same prompt format.

\subsubsection{Unconditional Replay Data Generation}
\label{app:replay-generate}

For each task $t>1$, the teacher generates $N_R$ sequences:
\begin{equation}
    \widetilde z_m
    =
    (\widetilde z_{m,1},\ldots,\widetilde z_{m,L_m})
    \sim
    p_{\bar{\Theta}_{t-1}}(\,\cdot\mid s),
    \qquad
    m=1,\ldots,N_R,
    \label{eq:app-replay-generate}
\end{equation}
where $\widetilde z_m$ is the $m$th generated sequence and $L_m$ is its length. We use nucleus sampling with top-$p=0.9$ and generation temperature $\tau_G$. Generation stops at the first end-of-sequence token or after $L_{\max}$ new tokens. In our experiments, we use $N_R=300$, $\tau_G=1.5$, and $L_{\max}=384$.

Each generation starts from the same one-token prompt $s$, with no query, answer, task index, or other information from an earlier example. Random sampling provides variation across generations. We call this replay unconditional because generation is not conditioned on an example or task identifier.

We discard empty continuations and collect the remaining samples in the temporary replay set
\begin{equation}
    \widetilde{\mathcal{D}}_t
    =
    \left\{
      (s,\widetilde z_m)
      :
      m=1,\ldots,N_R,
      \;L_m>0
    \right\}.
    \label{eq:app-replay-buffer}
\end{equation}
Thus $|\widetilde{\mathcal{D}}_t|\leq N_R$.

\subsubsection{Soft replay loss}
\label{app:replay-loss}

For each retained continuation, index the seed by $\ell=0$, the continuation tokens by $\ell=1,\ldots,L_m$, and any padding positions by $\ell>L_m$. Define the replay mask
\begin{equation}
    a_{m,\ell}
    =
    \mathbf{1}\!\left[1\leq \ell\leq L_m\right].
    \label{eq:app-replay-mask}
\end{equation}
At position $\ell$, the teacher and student receive the same prefix $(s,\widetilde z_{m,<\ell})$. Let
\begin{equation}
    \bar h_{m,\ell}
    =
    h_{\bar{\Theta}_{t-1}}(s,\widetilde z_{m,<\ell}),
    \qquad
    h_{m,\ell}
    =
    h_{\Theta}(s,\widetilde z_{m,<\ell}),
    \label{eq:app-replay-logits}
\end{equation}
denote their logits for the next token. For replay temperature $\tau_D$, define
\begin{equation}
    q_{m,\ell}^{D}
    =
    \operatorname{softmax}\!\left(\bar h_{m,\ell}/\tau_D\right),
    \qquad
    p_{m,\ell}^{D}
    =
    \operatorname{softmax}\!\left(h_{m,\ell}/\tau_D\right).
    \label{eq:app-replay-distributions}
\end{equation}
The data anchor minimizes the mask-normalized forward KL over the full vocabulary:
\begin{equation}
    \mathcal{L}_{D}^t
    =
    \frac{\tau_D^2}{\sum_{m,\ell}a_{m,\ell}}
    \sum_{m,\ell}
    a_{m,\ell}
    \KL\!\left(q_{m,\ell}^{D}\,\|\,p_{m,\ell}^{D}\right).
    \label{eq:app-replay-loss}
\end{equation}
The factor $\tau_D^2$ keeps the gradient scale comparable across replay temperatures.

The previous model determines both the replay sequences and the distributions that the student matches. At each generated-token position, the sampled continuation supplies a common prefix to the teacher and student, and $q_{m,\ell}^{D}$ is the teacher's next-token distribution. This soft target retains the teacher's uncertainty instead of treating the sampled token as the only correct outcome. The loss compares the full next-token distributions of the teacher and student.

When a configuration uses replay at task $t>1$, Equation~\ref{eq:app-fit} becomes
\begin{equation}
    \mathcal{L}_{\mathrm{fit}}^t
    =
    (1-w)\mathcal{L}_{\mathrm{SFT}}^t
    +w\mathcal{L}_{D}^t,
    \qquad
    w\in[0,1].
    \label{eq:app-replay-fit}
\end{equation}
The coefficients sum to one, so $w$ directly controls the balance between fitting the current task and matching the previous model on generated sequences. Increasing $w$ increases the replay contribution and decreases the current task contribution by the same amount.

\subsubsection{Training procedure}
\label{app:replay-procedure}

Algorithm~\ref{alg:replay} summarizes training for one task. Each epoch begins with a shuffled replay iterator. For every current minibatch, we draw one replay minibatch and create another shuffled iterator if the current one is exhausted. The current and replay losses therefore appear in the same number of minibatch updates. When the function anchor is active, we evaluate it only on the current minibatch.

\begin{algorithm}[H]
\DontPrintSemicolon
\KwIn{task data $\mathcal{D}_t$; student $\Theta$; previous model $\bar{\Theta}_{t-1}$ when $t>1$; seed $s$; attempts $N_R$; generation temperature $\tau_G$; maximum length $L_{\max}$; replay temperature $\tau_D$; replay weight $w$; epochs $E$}
\KwOut{updated student parameters}
\eIf{$t=1$}{
    train on $\mathcal{L}_{\mathrm{SFT}}^1$ \tcp*{$a_D^1=0$}
}{
    freeze $\bar{\Theta}_{t-1}$\;
    $\widetilde{\mathcal{D}}_t\gets\emptyset$\;
    \For{$m\gets1$ \KwTo $N_R$}{
        sample $\widetilde z_m\sim p_{\bar{\Theta}_{t-1}}(\cdot\mid s)$ with temperature $\tau_G$, top-$p=0.9$, and limit $L_{\max}$\;
        \If{$\widetilde z_m\neq\varnothing$}{
            add $(s,\widetilde z_m)$ to $\widetilde{\mathcal{D}}_t$\;
        }
    }
    \For{$e\gets1$ \KwTo $E$}{
        reshuffle the replay iterator\;
        \ForAll{current minibatches $\mathcal{B}\subset\mathcal{D}_t$}{
            draw replay minibatch $\widetilde{\mathcal{B}}\subset\widetilde{\mathcal{D}}_t$ \tcp*{recycle the iterator if needed}
            $\mathcal{L}\gets(1-w)\mathcal{L}_{\mathrm{SFT}}^t(\Theta,\mathcal{B})+w\mathcal{L}_{D}^t(\Theta,\widetilde{\mathcal{B}})$\;
            add the active function, weight, and low-rank allocation terms from Equation~\ref{eq:app-objective}\;
            take one optimizer step\;
        }
    }
    discard $\widetilde{\mathcal{D}}_t$ and release $\bar{\Theta}_{t-1}$\;
}
\caption{Unconditional generative replay for task $t$. The previous model generates the replay set before the student receives its first update on task $t$.}
\label{alg:replay}
\end{algorithm}

\subsubsection{Relation to prior formulations}
\label{app:replay-prior}

Deep Generative Replay introduced a generator and solver construction in which generated examples replace unavailable earlier data \citep{shin2017continual}. LAMOL uses one autoregressive language model to solve tasks and generate pseudo-samples. It trains each example in separate question-answering and generation formats, and studies both a shared generation token and task-specific generation tokens \citep{sun2019lamol}. Our data anchor uses one shared token in its training and evaluation format and never supplies task identity during generation. It also matches the teacher's full next-token distributions instead of maximizing the likelihood of sampled replay tokens.

\subsection{Function anchor: previous-state self-distillation}
\label{app:function-anchor}

The function anchor uses standard forward-KL distillation to limit changes in the model's output distributions. Inspired by Learning without Forgetting, which evaluates a previous model on current task inputs when earlier inputs are unavailable \citep{li2017learning} we apply this idea by comparing distributions over the full vocabulary of LLMs at each token position.

At the start of task $t>1$, we freeze the model obtained at task $t-1$ as $\bar{\Theta}_{t-1}$. For a current minibatch $\mathcal{B}$, let $n$ index a sequence $z_n=(z_{n,1},\ldots,z_{n,L_n})$, and let $\ell=1,\ldots,L_{n-1}$ index the prefix that predicts $z_{n,\ell+1}$. We write $h_{n,\ell}=h_{\Theta}(z_{n,\leq\ell})$ for the student logits and $\bar h_{n,\ell}=h_{\bar{\Theta}_{t-1}}(z_{n,\leq\ell})$ for the teacher logits. At temperature $\tau_F$, define
\begin{equation}
    q_{n,\ell}^{F}
    =
    \operatorname{softmax}(\bar h_{n,\ell}/\tau_F),
    \qquad
    p_{n,\ell}^{F}
    =
    \operatorname{softmax}(h_{n,\ell}/\tau_F).
    \label{eq:app-function-distributions}
\end{equation}
Here $q_{n,\ell}^{F}$ is the teacher distribution and $p_{n,\ell}^{F}$ is the student distribution. Using the sequence mask $m_{n,\ell+1}$ from Equation~\ref{eq:app-sft}, we average the forward KL over every nonpadding target position across the full vocabulary:
\begin{equation}
    \mathcal{L}_{F}^t
    =
    \frac{\tau_F^2}{\sum_{n,\ell}m_{n,\ell+1}}
    \sum_{n,\ell}
    m_{n,\ell+1}
    \KL\!\left(q_{n,\ell}^{F}\,\|\,p_{n,\ell}^{F}\right).
    \label{eq:app-function-loss}
\end{equation}
At each task boundary, we replace the teacher with the model that has just completed training. The teacher therefore is the model trained on task $t-1$. Shared LoRA freezes a copy of the previous adapter. Merged LoRA and sequential OSRM use the committed dense weights with a zero residual from the new adapter. O-LoRA uses the accumulated frozen factors and holds the new $B_t$ at zero during teacher evaluations. We run the teacher in evaluation mode, compute no gradients through it, and release it after the task.

The function and data anchors differ in the inputs on which they compare teacher and student. Equation~\ref{eq:app-function-loss} uses only sequences from the current task and generates no additional inputs. The data anchor instead samples sequences from the previous model and evaluates its loss on those generated sequences. When a configuration uses both anchors, they may share the same frozen teacher, but each anchor evaluates it on its own input set.

\subsection{Weight anchors}
\label{app:weight-anchors}

A weight anchor limits changes to trainable coordinates that earlier tasks marked as important. Let $\vartheta=(\vartheta_1,\ldots,\vartheta_P)$ denotes the $P$ tracked scalar coordinates. In our experiments, each coordinate is one scalar entry of a trainable LoRA factor. The implementation identifies each coordinate by its parameter name and position within the tensor. Both weight anchors use the following diagonal quadratic penalty:
\begin{equation}
    \mathcal{R}^{t+1}
    =
    \sum_{i=1}^{P}
    \Omega_{t,i}
    \left(
      \vartheta_i-\vartheta_{t,i}^{\star}
    \right)^2,
    \label{eq:app-weight-anchor-form}
\end{equation}
where $\vartheta_{t,i}^{\star}$ is the reference value stored after task $t$ and $\Omega_{t,i}$ measures the importance of coordinate $i$. Synaptic Intelligence (SI) estimates this importance from the optimization path during training. Online EWC estimates it from gradients of the trained model after training on the task. In the online EWC penalty, the running Fisher value $\widetilde F_{t,i}$ takes the place of $\Omega_{t,i}$ in Equation~\ref{eq:app-weight-anchor-form}.

The general quadratic in Equation~\ref{eq:weight-anchor} matches the implemented penalties by setting $H=2\operatorname{diag}(\Omega)$ for SI and $H=2\operatorname{diag}(\widetilde F)$ for online EWC. The coefficients $\lambda_{\mathrm{SI}}$ and $\lambda_{\mathrm{EWC}}$ absorb this constant factor. Section~\ref{app:coordinate-compat} explains the consequences of attaching each importance value to a named LoRA coordinate.

\subsubsection{Synaptic Intelligence}
\label{app:si}

SI assigns importance according to how much each coordinate contributed to reducing the fit loss during optimization \citep{zenke2017continual}. Let $K_t$ be the number of optimizer steps on task $t$, and let $\vartheta_{t,k,i}$ be coordinate $i$ immediately before step $k$. Thus $\vartheta_{t,0,i}$ is its value at the start of the task and $\vartheta_{t,K_t,i}$ is its value after the final step. For each optimizer step, SI accumulates the gradients from the microbatches in that step and records
\begin{equation}
    g_{t,k,i}
    =
    \frac{\partial\mathcal{L}_{\mathrm{fit}}^t}
         {\partial\vartheta_{t,k,i}},
    \label{eq:app-si-gradient}
\end{equation}
the gradient of the fit term with respect to coordinate $i$. The implementation computes this gradient before adding the function and weight penalties. It then captures the unclipped gradient, clips the complete-objective gradient in Equation~\ref{eq:app-objective}, and updates the parameters. SI therefore pairs an unclipped fit gradient with the parameter change, which can reflect every active objective term and gradient clipping.

SI accumulates the coordinatewise path contribution over task $t$:
\begin{equation}
    \omega_{t,i}
    =
    -\sum_{k=0}^{K_t-1}
    g_{t,k,i}
    \left(
      \vartheta_{t,k+1,i}-\vartheta_{t,k,i}
    \right).
    \label{eq:app-si-path}
\end{equation}
The summand is positive when a coordinate moves against its gradient diection, which is the local direction that reduces the loss. Thus $\omega_{t,i}$ gives a first order estimate of how much coordinate $i$ contributed to reducing the fit loss along the optimization path.

Let
\begin{equation}
    \Delta_{t,i}
    =
    \vartheta_{t,K_t,i}-\vartheta_{t,0,i}
    \label{eq:app-si-displacement}
\end{equation}
denote the total displacement of coordinate $i$ after we train the model on task $t$. Starting from $\Omega_{0,i}=0$, SI updates the cumulative importance and the reference before we start training on the next task:
\begin{equation}
    \Omega_{t,i}
    =
    \Omega_{t-1,i}
    +
    \frac{\max\!\left(0,\omega_{t,i}\right)}
         {\Delta_{t,i}^{2}+\xi},
    \qquad
    \vartheta_{t,i}^{\star}
    =
    \vartheta_{t,K_t,i}.
    \label{eq:app-si-consolidate}
\end{equation}
Dividing by $\Delta_{t,i}^2$ gives more importance to a coordinate that reduced the loss with less movement. The constant $\xi>0$ keeps the estimate stable when the total displacement is close to zero. Clamping the estimate at zero prevents the current task from assigning negative importance and reducing the penalty carried over from earlier tasks. SI combines the importance estimates from all completed tasks and uses the latest parameter values as the reference. It stores no task-specific states, so its retained state size does not grow with the number of tasks.

Starting with task $t+1$, SI uses
\begin{equation}
    \mathcal{R}_{\mathrm{SI}}^{t+1}
    =
    \sum_{i=1}^{P}
    \Omega_{t,i}
    \left(
      \vartheta_i-\vartheta_{t,i}^{\star}
    \right)^2.
    \label{eq:app-si-penalty}
\end{equation}
 When the data anchor is active, this term contains both the current task loss and the replay loss in Equation~\ref{eq:app-fit}. SI therefore attributes the update to both sources. O-LoRA is the exception: its SI state records only the weighted current data loss, although the optimizer also uses the replay loss and the active regularizers. O-LoRA also resets its trainable coordinate frame at each task boundary. Section~\ref{app:coordinate-compat} discusses the effect of this reset.

\subsubsection{Online EWC}
\label{app:online-ewc}

EWC uses Fisher information to measure how strongly the trained model's predictions depend on each coordinate \citep{kirkpatrick2017overcoming}. We use the online update from \citet{schwarz2018progress}. It combines information from completed tasks in one running Fisher, so its retained state size does not grow with the number of tasks.

At the end of task $t$, let $N_t=|\mathcal{D}_t|$ be the number of training sequences, let $N_{\mathrm{F}}$ be the maximum number of sequences used for Fisher estimation, and let $M_t=\min(N_t,N_{\mathrm{F}})$. After training on task $t$, we estimate the diagonal Fisher using the first $M_t$ sequences. For each sequence, we compute its causal language modeling loss with the model in evaluation mode and gradients enabled. Let $\ell_{t,n}$ denote the loss for sequence $n$. The diagonal Fisher estimate for task $t$ is
\begin{equation}
    F_{t,i}
    =
    \frac{1}{M_t}
    \sum_{n=1}^{M_t}
    \left(
      \frac{\partial\ell_{t,n}}
           {\partial\vartheta_i}
    \right)^2.
    \label{eq:app-fisher}
\end{equation}

Task Fishers can differ greatly in overall magnitude. Following \citet{schwarz2018progress}, we normalize each task Fisher before adding it to the running estimate so that relative parameter importance, rather than raw Fisher magnitude, determines its contribution. We use the first task's mean Fisher as the common scale. Define
\begin{equation}
    \bar F_t
    =
    \frac{1}{P}\sum_{i=1}^{P}F_{t,i},
    \qquad
    \widehat F_{t,i}
    =
    F_{t,i}\frac{\bar F_1}{\bar F_t},
    \label{eq:app-fisher-normalize}
\end{equation}
where $\bar F_t$ is the mean Fisher for task $t$ and $\bar F_1$ fixes the reference scale. This rescaling preserves the relative importance of coordinates within each task while keeping all task Fishers on the raw scale established by the first task. Scaling each Fisher to unit mean would instead change the effective strength of $\lambda_{\mathrm{EWC}}$ by several orders of magnitude in our setting.

Starting from $\widetilde F_{0,i}=0$, online EWC uses a decay coefficient $\gamma\in[0,1]$ and updates the running Fisher and the reference as
\begin{equation}
    \widetilde F_{t,i}
    =
    \gamma\widetilde F_{t-1,i}
    +\widehat F_{t,i},
    \qquad
    \vartheta_{t,i}^{\star}
    =
    \vartheta_{t,K_t,i}.
    \label{eq:app-online-ewc-update}
\end{equation}
Task $t+1$ then uses
\begin{equation}
    \mathcal{R}_{\mathrm{EWC}}^{t+1}
    =
    \sum_{i=1}^{P}
    \widetilde F_{t,i}
    \left(
      \vartheta_i-\vartheta_{t,i}^{\star}
    \right)^2.
    \label{eq:app-online-ewc-penalty}
\end{equation}
The decay coefficient $\gamma$ appears only in Equation~\ref{eq:app-online-ewc-update} because the running Fisher already contains the decay. Applying $\gamma$ again in the penalty would reduce earlier contributions twice. Our experiments use $\gamma=1$, so the running Fisher becomes a sum of the normalized task Fishers and retains all earlier contributions.

\subsubsection{Training procedure}
\label{app:weight-anchor-procedure}

Algorithm~\ref{alg:weight-anchors} shows when the two anchors collect and update their state. SI records information during ordinary training and requires no additional data pass. Online EWC estimates its Fisher after training and requires one backward pass for each selected sequence. Both methods consolidate the trained parameters before a LoRA merge or an O-LoRA fold changes the current factors. This order ensures that the new importance values and reference describe the coordinates used to learn task $t$.

\begin{algorithm}[H]
\DontPrintSemicolon
\KwIn{task data $\mathcal{D}_t$; state $(\Omega_{t-1},\vartheta_{t-1}^{\star})$ or $(\widetilde F_{t-1},\vartheta_{t-1}^{\star})$; damping $\xi$; decay $\gamma$; Fisher budget $N_{\mathrm{F}}$}
\KwOut{updated anchor state}
snapshot $\vartheta_{t,0}$; set $\omega_t\gets0$ \tcp*{SI only}
\For{$k\gets0$ \KwTo $K_t-1$}{
    accumulate $g_{t,k}$ from the fit term before adding penalties\;
    backpropagate the complete objective over the same accumulation window\;
    adjust an incomplete final window; capture $g_{t,k}$; clip the complete-objective gradient\;
    take one optimizer step\;
    $\omega_t\gets\omega_t-g_{t,k}\odot(\vartheta_{t,k+1}-\vartheta_{t,k})$ \tcp*{SI only}
}
\uIf{SI}{
    $\Omega_t\gets\Omega_{t-1}+\max(0,\omega_t)\oslash(\Delta_t^{\odot2}+\xi)$\;
}
\uElseIf{online EWC}{
    estimate $F_t$ from $\min(N_t,N_{\mathrm{F}})$ individual sequences\;
    rescale $F_t$ to $\widehat F_t$\;
    $\widetilde F_t\gets\gamma\widetilde F_{t-1}+\widehat F_t$\;
}
$\vartheta_t^{\star}\gets\vartheta_{t,K_t}$ \tcp*{before a merge or fold}
\caption{Weight-anchor update at task $t$. SI forms importance from the optimization path, while online EWC forms it from gradients of the trained model.}
\label{alg:weight-anchors}
\end{algorithm}

\subsubsection{Coordinate dependence of weight anchors}
\label{app:coordinate-compat}

After each task, SI and online EWC save the current LoRA parameters as a reference and estimate how important each parameter is for retaining what the model has learned. Shared LoRA keeps the same $A$ and $B$ matrices across tasks, so Equations~\ref{eq:app-si-penalty} and \ref{eq:app-online-ewc-penalty} compare each entry with its earlier value. Merged LoRA instead adds the completed update $B_tA_t$ to the dense weight matrix and creates new $A$ and $B$ matrices with the same names and dimensions. Sequential OSRM follows the same procedure but uses its own rule to initialize $A$. O-LoRA moves the completed update into its frozen accumulated matrices and then reinitializes the trainable $A$ and $B$ matrices without changing their names. After these resets, our implementation matches each stored reference and importance weight to the entry at the same row and column in the new matrices. An old importance weight therefore applies to a newly initialized parameter, even though changing that parameter may now affect the dense weight differently. This procedure defines a numerical penalty, but it does not preserve the original meaning of the importance weights across a reset.

To preserve the previous penalty after changing the parameter representation, we would need to transform both its reference values and its importance matrix. Suppose an invertible map $\vartheta_{\mathrm{old}}=f_t(\vartheta_{\mathrm{new}})$ relates the two representations. The new reference $\vartheta_{\mathrm{new}}^\star$ must satisfy $f_t(\vartheta_{\mathrm{new}}^\star)=\vartheta_{\mathrm{old}}^\star$. Let $J_t$ denote the Jacobian of $f_t$ at $\vartheta_{\mathrm{new}}^\star$. For small changes around the two reference points, preserving the quadratic penalty requires
\begin{equation}
    \delta\vartheta_{\mathrm{old}}
    =
    J_t\delta\vartheta_{\mathrm{new}},
    \qquad
    H_{\mathrm{new}}
    =
    J_t^\top H_{\mathrm{old}}J_t,
    \label{eq:transport}
\end{equation}
where $H_{\mathrm{old}}$ and $H_{\mathrm{new}}$ contain the importance weights in the two representations. This transformation is exact when $f_t$ is linear and describes the local behavior near the reference when $f_t$ is nonlinear. Even if $H_{\mathrm{old}}$ is diagonal, $H_{\mathrm{new}}$ generally contains off-diagonal terms that connect changes in different parameters. SI and online EWC store only one importance weight per parameter, so they cannot fully represent these interactions. When a low-rank allocation rule resets the LoRA workspace, it initializes new $A$ and $B$ matrices. Our implementation keeps the SI or online EWC reference and importance tensors and matches each stored value to the parameter at the same row and column in the new matrices. Since we don't have a transformation defined above, we do not adjust these stored values for the change in $A$ and $B$. Thus, the penalties applied to the new LoRA matrices may not have the same effect when they are applied to a dense model.

\subsection{Low-Rank Allocation Rules}
\label{app:allocation-details}

Let $\mathcal{J}$ denote the set of linear transformations to which we apply LoRA. For target $j\in\mathcal{J}$, let $W_{0,j}\in\mathbb{R}^{d_{\mathrm{out},j}\times d_{\mathrm{in},j}}$ be the pretrained weight matrix, and let $A_{t,j}\in\mathbb{R}^{r\times d_{\mathrm{in},j}}$ and $B_{t,j}\in\mathbb{R}^{d_{\mathrm{out},j}\times r}$ denote the LoRA matrices for module $j$ during task $t$, with LoRA rank $r$. The LoRA scale is $\rho=\alpha_{\mathrm{LoRA}}/r$. We write $(A_{t,j}^{(0)},B_{t,j}^{(0)})$ for the LoRA matrices at the start of task $t$ and $(A_{t,j}^{\star},B_{t,j}^{\star})$ for their values after training on that task. The matrix $W_{t,j}^{\mathrm{eff}}$ denotes the effective dense weight of module $j$ after task $t$, obtained by combining the base weight with all LoRA updates active in evaluation mode. The low-rank allocation rules differ in whether they reuse, merge, or retain the task factors.

\subsubsection{Shared LoRA}
\label{app:shared-allocation}

Shared LoRA uses one LoRA while we train it over all the tasks \citep{hu2022lora}. Let $\operatorname{Opt}_t$ denote all AdamW updates performed on task $t$, including the task-specific learning-rate schedule and every active term in Equation~\ref{eq:app-objective}. During task $t$,
\begin{equation}
    W_{t,j}^{\mathrm{eff}}
    =
    W_{0,j}
    +\rho B_{t,j}A_{t,j},
    \qquad
    (A_{t,j}^{\star},B_{t,j}^{\star})
    =
    \operatorname{Opt}_t
    \left(A_{t,j}^{(0)},B_{t,j}^{(0)}\right).
    \label{eq:app-shared}
\end{equation}
For the first task, the implementation initializes $A_{1,j}^{(0)}$ with Kaiming initialization and sets $B_{1,j}^{(0)}=0$. For each later task, it sets $(A_{t,j}^{(0)},B_{t,j}^{(0)})=(A_{t-1,j}^{\star},B_{t-1,j}^{\star})$. It creates a new optimizer and learning-rate schedule at every task boundary but keeps the learned factors. Thus, the cumulative adaptation of each target weight matrix has rank at most $r$, regardless of the number of tasks. Plain sequential supervised fine-tuning uses this low-rank allocation rule with every anchor indicator in Equation~\ref{eq:app-objective} set to zero.

\subsubsection{merged LoRA}
\label{app:merge-allocation}

Merged LoRA uses the merge and restart pattern associated with ReLoRA \citep{lialin2024relora}. ReLoRA schedules several restarts during pretraining and partially resets the optimizer state. This low-rank allocation rule instead performs one merge at each task boundary and creates a new optimizer for the next task.

Let $W_{t-1,j}$ denote the dense weight matrix obtained after merging the LoRA updates learned from tasks $1$ through $t-1$. Task $t$ uses a rank-$r$ LoRA:
\begin{equation}
    W_{t,j}^{\mathrm{eff}}
    =
    W_{t-1,j}
    +\rho B_{t,j}A_{t,j}.
    \label{eq:app-merge-forward}
\end{equation}
After training, the method folds the final task factors into the dense matrix:
\begin{equation}
    W_{t,j}
    =
    W_{t-1,j}
    +\rho B_{t,j}^{\star}A_{t,j}^{\star}.
    \label{eq:app-merge}
\end{equation}
The implementation then removes the old adapter and attaches a new one with $A_{t+1,j}^{(0)}$ initialized by the Kaiming rule and $B_{t+1,j}^{(0)}=0$. Because the new LoRA update is initialized to zero, $B_{t+1,j}A_{t+1,j}=0$, task $t+1$ starts from the function represented by the merged weight $W_{t,j}$. Each task contributes a matrix of rank at most $r$, so the sum of the committed residuals can reach rank $tr$.

\subsubsection{O-LoRA}
\label{app:olora}

O-LoRA assigns a new rank-$r$ LoRA to each task and keeps all earlier pairs fixed \citep{wang2023orthogonal}. Our implementation follows the released two-pair structure: one concatenated pair stores completed tasks, while one fixed-rank pair learns the current task. The original method applies LoRA to the query and value projections. For a controlled comparison, we apply every low-rank allocation rule to the common target set in Table~\ref{tab:implementation-hparams}.

At the start of task $t$, define the accumulated factors
\begin{equation}
    A_{<t,j}
    =
    \begin{bmatrix}
      A_{1,j}^{\star}\\[-1mm]\vdots\\A_{t-1,j}^{\star}
    \end{bmatrix},
    \qquad
    B_{<t,j}
    =
    \begin{bmatrix}
      B_{1,j}^{\star}&\cdots&B_{t-1,j}^{\star}
    \end{bmatrix}.
    \label{eq:app-olora-concat}
\end{equation}
The represented target matrix is
\begin{equation}
    W_{t,j}^{\mathrm{eff}}
    =
    W_{0,j}
    +\rho B_{<t,j}A_{<t,j}
    +\rho B_{t,j}A_{t,j}.
    \label{eq:app-olora-forward}
\end{equation}
Only $(A_{t,j},B_{t,j})$ receives gradients. The accumulated LoRAs remain frozen. At the start of the task, we initialize $A_{t,j}$ with the Kaiming rule and sets $B_{t,j}=0$.

O-LoRA encourages the current input-side factors to differ from the accumulated input-side factors. The implementation uses
\begin{equation}
    \mathcal{R}_{\perp}^t
    =
    \sum_{j\in\mathcal{J}}
    \left\|
      A_{<t,j}A_{t,j}^{\top}
    \right\|_{1,1},
    \qquad
    \|C\|_{1,1}=\sum_{a,b}|C_{a,b}|.
    \label{eq:app-olora-orth}
\end{equation}
This penalty is zero on the first task because no earlier LoRAs exist. The released implementation uses the entrywise $\ell_1$ norm in Equation~\ref{eq:app-olora-orth}, whereas the paper presents a squared Frobenius penalty. We follow the released implementation so the experiment matches its executable training rule.

The original code also provides the optional factor norm
\begin{equation}
    \mathcal{R}_{2}^t
    =
    \sum_{j\in\mathcal{J}}
    \left(
      \|A_{t,j}\|_{F}
      +\|B_{t,j}\|_{F}
    \right).
    \label{eq:app-olora-l2}
\end{equation}
The regular configuration sets $\lambda_2=0$, so this term does not affect training. Our implementation adds $\mathcal{R}_{\perp}^t$ and $\mathcal{R}_2^t$ without dividing them by the gradient-accumulation factor, following the released code.

After task $t$, the method appends $(A_{t,j}^{\star},B_{t,j}^{\star})$ to the accumulated LoRA set and start a new LoRA for training on the next task. The model after task $t$ therefore represents
\begin{equation}
    W_{t,j}
    =
    W_{0,j}
    +\rho\sum_{s=1}^{t}B_{s,j}^{\star}A_{s,j}^{\star}.
    \label{eq:app-olora-sum}
\end{equation}
The evaluation code reconstructs the same matrix by adding the saved rank-$r$ LoRA for tasks $1$ through $t$ to a fresh base model. It does not select adapters using a task identifier.

\subsubsection{Sequential OSRM adaptation}
\label{app:osrm}

OSRM chooses each LoRA input subspace from features of the other tasks before fine-tuning and then merges the independently trained task models \citep{zhang2025unraveling}. This procedure assumes access to training data from all tasks. Our sequential adaptation cannot use future tasks, so it uses features only from completed tasks to initialize the next LoRA workspace.

For a completed task $s$, let $N_s=|\mathcal{D}_s|$, let $N_H$ be the feature-sample budget, and let $M_s^H=\min(N_s,N_H)$. The implementation takes the first $M_s^H$ datapoints in the task data order and divides them into $C_s$ forward batches. For each LoRA-adapted linear module $j$, let $h_{j,\ell}(z)\in\mathbb{R}^{d_{\mathrm{in},j}}$ denote the input hidden-state vector at sequence position $\ell$, measured immediately before the module applies the LoRA down-projection matrix $A_j$ while processing the training sequence $z$. Let $\mathcal{B}_{s,b}$ be feature batch $b$ and let $L_{s,b}$ be its padded sequence length. The stored feature is
\begin{equation}
    \bar h_{s,j}
    =
    \frac{1}{C_s}
    \sum_{b=1}^{C_s}
    \frac{1}{|\mathcal{B}_{s,b}|L_{s,b}}
    \sum_{z\in\mathcal{B}_{s,b}}
    \sum_{\ell=1}^{L_{s,b}}
    h_{j,\ell}(z).
    \label{eq:app-osrm-feature}
\end{equation}
We first average over positions and sequences within each batch and then gives every batch mean equal weight. Before task $t$, the method stacks the feature vectors from completed tasks:
\begin{equation}
    H_{<t,j}
    =
    \begin{bmatrix}
      \bar h_{1,j}^{\top}\\[-1mm]\vdots\\
      \bar h_{t-1,j}^{\top}
    \end{bmatrix}
    \in\mathbb{R}^{(t-1)\times d_{\mathrm{in},j}}.
    \label{eq:app-osrm-stack}
\end{equation}
Let $H_{<t,j}=U\Sigma V^{\top}$ be its full singular value decomposition, and let $V_{\min}\in\mathbb{R}^{d_{\mathrm{in},j}\times r}$ contain the $r$ right singular vectors associated with the smallest singular values. Let $\widehat A_{t,j}$ denote the ordinary Kaiming initialization drawn when the implementation attaches the fresh adapter. Define
\begin{equation}
    \kappa_{t,j}
    =
    \frac{1}{r}
    \sum_{a=1}^{r}
    \left\|\widehat A_{t,j}[a,:]\right\|_2,
    \qquad
    A_{t,j}^{(0)}
    =
    \kappa_{t,j}V_{\min}^{\top},
    \qquad
    B_{t,j}^{(0)}=0.
    \label{eq:app-osrm-init}
\end{equation}
The smallest right singular vectors identify input directions with the least energy in the stored past-task means. When the null space of $H_{<t,j}$ has dimension at least $r$, the selected rows lie in that null space and satisfy $A_{t,j}^{(0)}\bar h_{s,j}=0$ for every stored task $s<t$. The factor $\kappa_{t,j}$ restores the mean row norm of the ordinary Kaiming initialization because unscaled singular vectors have unit norm.

The first task uses ordinary LoRA initialization because it has no past features. After each later task, the implementation first collects $\bar h_{t,j}$ from the trained model, then merges the learned LoRA matrices with Equation~\ref{eq:app-merge}, and finally initializes the next fresh LoRA from all stored feature means. Relative to the original post-training merge setting, this version makes three changes: it uses only past features, extracts them from the just-trained model rather than a common pretrained model, and rescales the selected directions before optimizing both factors. Thus, we refer to it as a sequential OSRM adaptation.

\subsection{Training Procedure}
\label{app:boundary-order}

Algorithm~\ref{alg:app-task-boundary} records the training procedure. The order of combining different algorithms matters because teachers must describe the model before task $t$, while the weight anchors and OSRM features must describe the model after it learns task $t$ but before a low-rank allocation rule changes its coordinates.

\begin{algorithm}[H]
\DontPrintSemicolon
\KwIn{task data $\mathcal{D}_t$; learner state after task $t-1$; active anchors; low-rank allocation rule}
\KwOut{learner state after task $t$}
load and format $\mathcal{D}_t$\;
\If{$t>1$}{
    construct each active previous-state teacher\;
    generate the fixed-budget replay set if the data anchor is active\;
}
snapshot the trainable starting coordinates if SI is active\;
optimize Equation~\ref{eq:app-objective} on task $t$\;
consolidate SI if it is active; estimate the online-EWC Fisher if it is active\;
collect $\{\bar h_{t,j}:j\in\mathcal{J}\}$ from the trained model if sequential OSRM is active\;
\uIf{shared LoRA}{
    retain $(A_{t,j}^{\star},B_{t,j}^{\star})$ for task $t+1$\;
}
\uElseIf{merged LoRA}{
    commit Equation~\ref{eq:app-merge}; attach a fresh rank-$r$ workspace\;
}
\uElseIf{O-LoRA}{
    save the current rank-$r$ factors; append them to the frozen accumulated pair; reset the workspace\;
}
\Else{
    commit Equation~\ref{eq:app-merge}; attach a fresh workspace; apply Equation~\ref{eq:app-osrm-init} \tcp*{sequential OSRM}
}
\caption{Order of the operations at the boundary of task $t$.}
\label{alg:app-task-boundary}
\end{algorithm}

To compare persistent method state, let
\begin{equation}
    P_r
    =
    \sum_{j\in\mathcal{J}}
    r\left(d_{\mathrm{in},j}+d_{\mathrm{out},j}\right)
    \label{eq:app-lora-storage}
\end{equation}
be the number of scalars in one rank-$r$ adapter. This count excludes evaluation checkpoints and diagnostic logs because they do not form part of the state used to learn later tasks. Shared LoRA stores one $P_r$ LoRA adapter. Merged LoRA stores one dense model and one $P_r$ LoRA adapter, so its additional adapter state does not grow with the number of completed tasks. After $t$ completed tasks, O-LoRA stores $tP_r$ accumulated scalars and one $P_r$ LoRA adapter while training the next task. Sequential OSRM stores the merged LoRA state together with one $d_{\mathrm{in},j}$-dimensional feature vector for every completed task and target, so its retained feature state has size $O(t\sum_{j\in\mathcal{J}}d_{\mathrm{in},j})$.

\subsection{Optimization and Method Settings}
\label{app:hyperparameters}
Table~\ref{tab:implementation-hparams} records the settings used by the registered canonical runs. For hyperparameters selected by task-level successive halving, the table gives both the candidate set and the registered value.

\begin{table}[H]
    \centering
    \small
    \begin{tabular}{p{0.24\linewidth}p{0.67\linewidth}}
        \hline
        Component & Setting \\
        \hline
        Backbone & Qwen3-4B-Base \\
        Current-task optimization & 10 epochs per task; minibatch size 8; gradient accumulation 1; maximum sequence length 384 \\
        Optimizer & AdamW; learning rate $5\times10^{-4}$; weight decay $0.01$; global gradient-norm clipping at 1 \\
        Per-task schedule & Linear warmup during the first $5\%$ of optimizer steps, followed by a constant learning rate; optimizer and schedule restart for every task \\
        LoRA & $r=32$; $\alpha_{\mathrm{LoRA}}=64$; $\rho=2$; dropout $0.05$; target modules $\{$\texttt{q\_proj}, \texttt{k\_proj}, \texttt{v\_proj}, \texttt{o\_proj}, \texttt{gate\_proj}, \texttt{up\_proj}, \texttt{down\_proj}$\}$ \\
        Data anchor & $N_R=300$; $\tau_D=2$; frozen replay seed token; top-$p=0.9$; generation batch size 32; maximum 384 new tokens; generation-temperature candidates $\{1.0,1.5\}$ and registered value $1.5$; $w$ candidates $\{0.5,0.75\}$ and registered values $0.75$ (\symbolqa{}), $0.5$ (\llmqa{}), and $0.75$ (\realqa{}) \\
        Function anchor & Forward KL over the complete vocabulary; $\tau_F=5$; $\lambda_F$ candidates $\{1,3\}$ and registered value 1; previous-state teacher refreshed at every task \\
        SI & $\lambda_{\mathrm{SI}}=1$; $\xi=0.1$; negative task contributions clamped to zero \\
        Online EWC & $\lambda_{\mathrm{EWC}}=1000$; $\gamma=1$; at most $N_{\mathrm{F}}=1000$ sequences per task; normalization to the first task's mean Fisher enabled \\
        O-LoRA & $\lambda_{\perp}=0.5$; $\lambda_2=0$; rank 32 added for each task \\
        Sequential OSRM & Up to $N_H=64$ training sequences are used to compute one mean input-feature vector per adapted layer and task. Padded positions are included. Each new $A$ is initialized from the smallest right singular vectors and rescaled to the mean row norm of its Kaiming initialization \\
        \hline
    \end{tabular}
    \caption{Canonical implementation and optimization settings.}
    \label{tab:implementation-hparams}
\end{table}

The canonical suite uses unconditional KL replay with a frozen replay-token embedding and forward-KL function distillation over the complete vocabulary. It does not use query-conditioned replay. It also leaves the optional replay cross-entropy, trainable replay-token embedding, reverse KL, Jensen-Shannon divergence, vocabulary truncation, O-LoRA gradient projection, and O-LoRA factor-norm paths disabled. We therefore do not treat these options as separate methods in Table~\ref{tab:method-taxonomy}.

\section{Data Construction}
\label{app:data}
This section documents how we build three datasets for Section~\ref{sec:evaluation}. All three share the same interface: a dataset consists of 100 tasks, each holding a list of texts \texttt{Question: <q>\textbackslash nAnswer: <a>} whose test split is a copy of its train split. At training and evaluation time the answer is rendered as \texttt{\string\boxed\{a\}}, so the delimiter is learned as a format norm (Appendix~\ref{app:hyperparameters}).

\subsection{Data Uniqueness}
\label{app:data-uniqueness}

A checkpoint trained through task $i$ is queried with $z_{j,n}$ alone for all $j\le i$. If two tasks contained the same question with different answers, the target would be undefined and forgetting would be confounded with ambiguity. For the synthetically generated datasets, each generator therefore maintains a registry of emitted question keys and rejects any candidate whose key has already appeared. \symbolqa{} constrains keys only, since the key is what the model is queried on. \llmqa{} constrains both the invented entity and the full question string.

\subsection{\symbolqa{}}
\label{app:data-symbol}
\symbolqa{} contains $10{,}000$ random key--value associations divided evenly across 100 tasks. Each task contains 100 items. For each item, the generator samples a six-character key and a four-character value from the same 62-symbol alphabet of uppercase letters, lowercase letters, and digits. It redraws any key that has already appeared, ensuring that every key maps to a single value across the full dataset. All tasks use the same alphabet, lengths, and question format, so an item's surface form does not reveal its task. Because the generator selects each value independently of its key, there are no general patterns associating the keys to their answers. The model must therefore remember each key-- value association. Training and evaluation use the same associations, so this dataset measures retention rather than generalization to unseen pairs.

\subsection{\llmqa{}}
\label{app:data-llm}

\paragraph{Generation.} We use the instruction-tuned Qwen3-4B-Instruct-2507 model \citep{yang2025qwen3} to generate facts about fictional entities. The dataset defines 100 topics, such as ``fictional lighthouses and their keepers,'' and assigns one topic to each task. For every prompt, we ask the generator to produce ten candidate JSON records with the fields \texttt{entity}, \texttt{question}, and \texttt{answer}. The prompt instructs the model to invent entities and facts that do not exist and to avoid real people, places, organizations, works, and events.

\paragraph{Validation.} For each candidate, the validator extracts three non-empty fields and checks that the question contains the entity name after normalizing case and whitespace. This requirement makes the fictional entity the main information that distinguishes one question from another. The validator also limits answers to eight words and rejects questions or answers that contain line breaks. Across the full dataset, the validator rejects any candidate whose normalized entity or question matches one already accepted. Generation continues until every task contains 100 valid and globally distinct questions.

\paragraph{Task order.} The original topic list places related subjects near one another. After generation, we apply a fixed random permutation with seed 0 to break this ordering. The permutation changes only the task indices, and it leaves every entity, question, answer, and topic unchanged.

\paragraph{Novelty.} The generation prompt asks for fictional entities and facts, but the pipeline does not verify their nonexistence against an external source. We therefore do not claim that every \llmqa{} item was empirically unknown to the base model before training.

\subsection{\realqa{}}
\label{app:data-real}

\paragraph{Sources.}
We draw equally from ten public English question-answering datasets, five short-answer and five multiple-choice (Table~\ref{tab:realqa-sources}). Short-answer sources are TriviaQA \citep{joshi2017triviaqa}, the open-domain formulation \citep{lee2019latent} of Natural Questions, PopQA \citep{mallen2023not}, SQuAD \citep{rajpurkar2016squad}, and WebQuestions \citep{berant2013semantic}. Multiple-choice sources are OpenBookQA \citep{mihaylov2018can}, SciQ \citep{welbl2017crowdsourcing}, both partitions of ARC \citep{clark2018think}, and MedMCQA \citep{pal2022medmcqa}.

\paragraph{Normalization to free-form recall.}
Each source is mapped to a common record with a question, a primary answer, and an alias list. Multiple-choice items are converted by discarding the options and keeping the text of the correct choice as the target, so the model must produce the answer rather than select a letter. For datasets that consist of multiple correct answer aliases, we accept all of them as the correct answer.

\paragraph{Per-model contamination filter.}
Facts the base model already knows cannot be acquired, and would enter the accuracy matrix as spurious retention. For each candidate, we prompt the base model with \texttt{Question: <q>\textbackslash nAnswer:} and draw five samples (temperature $0.7$, top-$p=0.95$, at most 64 new tokens, stopping at a newline or a new question). A candidate is discarded if any of the five completions contains any accepted answer as a word-boundary substring, while the remainder are retained. Word boundaries matter because short aliases such as country codes otherwise match inside unrelated words. Retained items are therefore items the base model failed to produce under five attempts. We do not claim the complete absence of knowledge from the model's pretraining phase, but we treat this as an indication of the model's less familiar facts for it to learn in our training process.

\paragraph{Assembly.}
From each source we sample 500 filtered items with a fixed seed, giving $5000$ items. Sampling equally from ten sources rather than proportionally prevents the largest corpus from dominating the dataset. We then pool all $5000$ items, shuffle globally under a fixed seed, and cut the shuffled sequence into 100 consecutive tasks of 50. Each task is thus a uniform mixture of all ten sources. This procedure helps us to remove possible interference from dataset domains during training. For instance, if we don't do the shuffling, the model may learn from the same original dataset in a task and thus may have a better retention rate that is hard to explain.

\begin{table}[H]
    \centering
    \small
    \begin{tabular}{llll}
        \hline
        Source & Type & Original form & Retained \\
        \hline
        TriviaQA & short answer & trivia questions with alias sets & 500 \\
        NQ-Open & short answer & search queries, short answer list & 500 \\
        PopQA & short answer & entity-centric relational queries & 500 \\
        SQuAD & short answer & span answers over passages & 500 \\
        WebQuestions & short answer & Freebase question--answer pairs & 500 \\
        OpenBookQA & multiple choice & elementary science, four options & 500 \\
        SciQ & multiple choice & crowdsourced science questions & 500 \\
        ARC-Easy & multiple choice & grade-school science, four options & 500 \\
        ARC-Challenge & multiple choice & retrieval-resistant science questions & 500 \\
        MedMCQA & multiple choice & medical entrance-exam questions & 500 \\
        \hline
    \end{tabular}
    \caption{\realqa{} sources.}
    \label{tab:realqa-sources}
\end{table}

\subsection{What the three tiers isolate}
\label{app:data-comparison}

The three datasets test memorization with increasing amounts of familiar language and knowledge. All three use the same question and answer format, but they differ in how much the model can use what it learned before continual training. In \symbolqa{}, the key and value have no meaningful relationship, so the model must remember each individual association. \llmqa{} uses natural questions about invented entities. Familiar language and entity types may help the model encode these examples, but real-world knowledge does not determine their answers. \realqa{} uses questions and answers about real entities and filters out questions that the base model answers correctly before training. Comparing the three datasets shows whether a method behaves similarly for arbitrary associations, fictional facts expressed in natural language, and real knowledge.

\begin{table}[H]
    \centering
    \small
    \begin{tabular}{p{0.14\linewidth}p{0.13\linewidth}p{0.20\linewidth}p{0.22\linewidth}p{0.18\linewidth}}
        \hline
        Dataset & Tasks $\times$ items & Novelty control & Task organization & Semantic information \\
        \hline
        \symbolqa{} & $100\times100$ & Randomly generated symbol associations & One shared generation rule across all tasks & No structure in the target mapping \\
        \llmqa{} & $100\times100$ & Fictional generation prompt without a base model filter & One topic per task with a fixed random topic order & Natural language and familiar entity types \\
        \realqa{} & $100\times50$ & Five sample filter for each base model & Ten data sources mixed across tasks by a global shuffle & Natural language and real facts \\
        \hline
    \end{tabular}
    \caption{Comparison of the three memorization datasets. In every dataset, training and evaluation use the same questions, each question has one target answer across all tasks, and evaluation provides no task identifier.}
    \label{tab:dataset-summary}
\end{table}

\section{Search Procedure Details}
\label{app:search}

This section describes the task-level successive halving search introduced in Section~\ref{sec:search}. The search ranks 90 configurations that combine shared LoRA or merged LoRA with optional weight, function, and data anchors. O-LoRA and sequential OSRM do not enter this search pool because their stored state grows with the number of tasks. We evaluate them separately as task-growing low-rank allocation rules.

\begin{algorithm}[t]
\DontPrintSemicolon
\KwIn{candidate set $\mathcal{A}_1$, task horizons $r_1<\cdots<r_K$, survivor counts $n_2,\ldots,n_K$, seeds $\mathcal{S}$, fixed task order $\pi$}
\KwOut{candidate configurations ranked at horizon $r_K$}
\For{$k \gets 1$ \KwTo $K$}{
  \ForAll{$a\in\mathcal{A}_k$ and $s\in\mathcal{S}$}{
    \eIf{$k=1$}{
      train configuration $a$ from the pretrained model on tasks $1,\ldots,r_1$ using order $\pi$ and seed $s$\;
    }{
      restore the checkpoint and training state saved at horizon $r_{k-1}$, then train tasks $r_{k-1}{+}1,\ldots,r_k$\;
    }
    evaluate the resulting model on tasks $1,\ldots,r_k$ to obtain $M_{r_k,j}^{a,s}$ for $j\leq r_k$\;
  }
  score each $a\in\mathcal{A}_k$ by $F_{r_k}(a)$ from Equation~\ref{eq:search-score}\;
  rank $\mathcal{A}_k$ in descending order of $F_{r_k}(a)$\;
  \If{$k<K$}{
    $\mathcal{A}_{k+1}\gets$ the top $n_{k+1}$ configurations in $\mathcal{A}_k$\;
  }
}
\Return{the configurations in $\mathcal{A}_K$ in ranked order}\;
\caption{TSH over increasing task horizons. For each seed, a promoted configuration continues from its previous checkpoint instead of training the completed tasks again.}
\label{alg:search}
\end{algorithm}

\subsection{Cost Analysis}
\label{app:search-cost}
Let $r_k$ denote the cumulative number of tasks trained at search run $k$, and let $n_k=|\mathcal{A}_k|$ denote the number of configurations evaluated at that run. Let $n_k=|\mathcal{A}_k|$ denote the number of configurations that reach horizon $r_k$. We count training one configuration on one task as one unit of training cost. At run $k$, each of the $n_k$ configurations trains only the $r_k-r_{k-1}$ tasks added since the previous run. The total training cost per seed and dataset is therefore
\begin{equation}
    C
    =
    \sum_{k=1}^{K} n_k(r_k-r_{k-1}),
    \qquad
    r_0=0.
    \label{eq:app-search-cost}
\end{equation}
For the schedule in Section~\ref{sec:search},
\begin{equation}
    C
    =
    90(10)+45(10)+23(30)+10(50)
    =
    2540.
    \label{eq:app-search-cost-value}
\end{equation}
Training all 90 configurations for 100 tasks would instead require $9000$ such units per seed. The search therefore uses $28.2\%$ of the exhaustive training cost, which corresponds to a $71.8\%$ reduction.

\subsection{Selection and continuation}
\label{app:search-cuts}

Every configuration uses seeds 41, 42, and 43 at each horizon. We average the three scores before ranking, so one seed does not determine whether a configuration advances. After 10 tasks, the search retains the top $\lceil90/2\rceil=45$ configurations. After 20 tasks, it retains the top $\lceil45/2\rceil=23$. After 50 tasks, it retains the top 10 for the final 100 task horizon. The schedule starts at 10 tasks because pilot runs at five tasks provided little separation among the configurations.

When a configuration advances, the trainer restores its model parameters, the state required by its continual learning methods, and the random number generator states. It reconstructs the temporary teachers for the function and data anchors from the restored previous model at the start of the next task. With the same seed, task order, and training settings, this continuation reproduces the trajectory of an uninterrupted run rather than approximating it with a smaller model or less data.

\subsection{Implementation}
\label{app:search-impl}

For caching, we group the three seed runs for one configuration at one task budget into a search cell. Each cell records the model, dataset, task budget, method choices, method hyperparameters, fixed training settings, seed list, and task order. Before reusing a completed cell, the implementation checks that all recorded settings match the current search. This allows searches with different selection schedules to reuse previously computed results when the underlying experiments are identical. If any setting differs, the implementation aborts instead of reusing the cell. During the search, every configuration uses the task permutation generated with seed 1234. Because the cache record includes this task order, search results cannot be confused with results from the default order. After selection, we retrain the chosen configurations from the pretrained model on the default task order.

After evaluating a search cell, the implementation retains the checkpoint and continuation state saved after its last task and removes the earlier task checkpoints. These files contain everything needed to continue a promoted configuration at the next run.

\subsection{Search outputs}
\label{app:search-outputs}

For each dataset, the search writes a leaderboard and a trace containing every configuration's score, rank, and selection status at each horizon it reaches. We fix the selected configuration and its searched hyperparameters before running the final experiments. The final experiments retrain the selected configurations without further tuning and present the tasks in their numbered order for that dataset.

\subsection{Transfer to the report task order.}
\label{app:search-order}
The preceding analysis keeps the task order fixed. We next compare the search ranking at each stage with the 100-task ranking from the final experiments. The search uses the development task order, whereas the final experiments use a separate report task order. This comparison therefore tests whether the search ranking remains informative when both the number of tasks and their order change.

Search candidates vary in both method composition and hyperparameters, whereas the final experiments report one setting for each composition. To compare them, we group search candidates by their corresponding reported composition and represent each group by its highest-scoring hyperparameter variant at that search stage. This follows the search procedure, which promotes the best-performing variant. We include only compositions evaluated in the final experiments. Let $\Delta$ denote the difference between a composition's search rank and its final rank. Table~\ref{tab:rung-vs-canonical} reports the agreement between the two rankings.

After 10 tasks, all 17 comparable compositions remain in the search. The Spearman correlations with the final ranking are $0.93$, $0.95$, and $0.90$ on \symbolqa{}, \llmqa{}, and \realqa{}, respectively. The mean absolute rank displacement is $1.5$, $1.3$, and $1.4$ positions. Thus, the first search stage closely approximates the final ranking even though the final experiments use a different task order and continue training through 100 tasks.

We do not interpret the lower correlations at some later stages as evidence that the ranking becomes less reliable. Each search stage removes configurations, so the comparison set becomes smaller and increasingly consists of strong configurations with similar scores. A one-position change among these configurations can substantially change the correlation. For example, the 50-task comparison on \symbolqa{} contains only five compositions. We therefore use the 10-task results, which cover all 17 comparable compositions, as the primary check and report the later stages for completeness.

\begin{table}[H]
\centering
\small
\begin{tabular}{llrrrrr}
\hline
Dataset & Tasks & Compositions & Mean $|\Delta|$ & Max $|\Delta|$ & $\rho$ & Top-3 overlap \\
\hline
\symbolqa{} & 10 & 17 & 1.5 & 5 & 0.93 & 2/3 \\
 & 20 & 9 & 1.3 & 3 & 0.78 & 2/3 \\
 & 50 & 5 & 1.2 & 2 & 0.50 & 2/3 \\
 & 100 & 3 & 0.7 & 1 & 0.50 & 3/3 \\
\llmqa{} & 10 & 17 & 1.3 & 3 & 0.95 & 2/3 \\
 & 20 & 6 & 0.3 & 1 & 0.94 & 2/3 \\
 & 50 & 4 & 0.5 & 1 & 0.80 & 2/3 \\
 & 100 & 3 & 0.7 & 1 & 0.50 & 3/3 \\
\realqa{} & 10 & 17 & 1.4 & 5 & 0.90 & 3/3 \\
 & 20 & 8 & 0.0 & 0 & 1.00 & 3/3 \\
 & 50 & 4 & 0.0 & 0 & 1.00 & 3/3 \\
 & 100 & 3 & 0.0 & 0 & 1.00 & 3/3 \\
\hline
\end{tabular}
\caption{Agreement between the search ranking on the development task order and the final ranking on the report task order. We group search candidates by method composition and represent each composition by its highest-scoring hyperparameter variant at each search stage. The displacement $|\Delta|$ measures the absolute difference between the search and final ranks. Top-3 overlap reports how many of the three highest-ranked final compositions also appear in the top three at that search stage. The comparison set becomes smaller as the search removes candidates, so the 10-task rows provide the most complete comparison.}
\label{tab:rung-vs-canonical}
\end{table}

\section{Additional Results}
\label{app:results}

This section records everything the
main text compresses: every configuration on every dataset, the complete factorial including terms that are not significant, the task-growing low-rank allocation experiments, the compositions that fail outright, the seed-variance controls, and the held-out general capability evaluation.

\subsection{Cross-dataset ranking}
\label{app:cross-dataset-ranking}

Table~\ref{tab:app-factorial-ranks} ranks the 16 compositions in the main $2^4$ factorial by mean final retention within each dataset. For composition $m$, let $r_{m,d}$ be its rank on dataset $d$. We define $k_m=\max_d r_{m,d}$, which is the smallest $k$ for which the composition is among the top $k$ on every dataset. Lower values indicate more consistent performance across datasets. We use mean rank to order compositions with the same $k_m$. Our best method combines all three anchors with merged LoRA. It is the only composition that ranks among the top 3 methods in all datasets, so it uniquely attains $k_m=3$. Its average final retention is 34.9\% across the three datasets.

\begin{table}[H]
\centering
\small
\setlength{\tabcolsep}{5pt}
\begin{tabular*}{\linewidth}{@{\extracolsep{\fill}}lrrrrr@{}}
\toprule
Composition & \symbolqa{} & \llmqa{} & \realqa{} & $k_m$ & Mean rank \\
\midrule
\textbf{All anchors + merged LoRA} & \textbf{2} & \textbf{1} & \textbf{3} & \textbf{3} & \textbf{2.00} \\
SI + Replay + merged LoRA & 4 & 2 & 1 & 4 & 2.33 \\
SD + Replay + merged LoRA & 1 & 3 & 4 & 4 & 2.67 \\
Replay + merged LoRA & 3 & 4 & 2 & 4 & 3.00 \\
SI + SD + Replay & 5 & 5 & 8 & 8 & 6.00 \\
SD + merged LoRA & 6 & 8 & 6 & 8 & 6.67 \\
SI + Replay & 8 & 7 & 7 & 8 & 7.33 \\
SI + SD + merged LoRA & 10 & 6 & 5 & 10 & 7.00 \\
SI + SD & 9 & 10 & 9 & 10 & 9.33 \\
SD + Replay & 7 & 9 & 11 & 11 & 9.00 \\
Replay & 11 & 11 & 10 & 11 & 10.67 \\
SD & 12 & 13 & 12 & 13 & 12.33 \\
SI + merged LoRA & 14 & 14 & 13 & 14 & 13.67 \\
merged LoRA & 13 & 12 & 15 & 15 & 13.33 \\
SI & 15 & 15 & 14 & 15 & 14.67 \\
Naive fine-tuning & 16 & 16 & 16 & 16 & 16.00 \\
\bottomrule
\end{tabular*}
\caption{Cross-dataset ranks of the 16 compositions in the main factorial evaluation. Rows are ordered by the worst rank $k_m$, with mean rank used to break ties.}
\label{tab:app-factorial-ranks}
\end{table}

\subsection{Complete per-configuration results}
\label{app:res-all-configs}

\begin{table}[H]
\centering
\small
\begin{tabular}{lrrrrr}
\hline
configuration & Final & Diag & Forget & $W(10)$ & $W(50)$ \\
\hline
\texttt{sd\_replay\_merge} & 23.2 $\pm$ 4.2 & 99.5 $\pm$ 0.3 & 77.2 $\pm$ 4.1 & 95.4 $\pm$ 1.7 & 45.9 $\pm$ 8.1 \\
\texttt{si\_sd\_replay\_merge} & 18.5 $\pm$ 2.8 & 99.3 $\pm$ 0.1 & 81.8 $\pm$ 2.8 & 93.9 $\pm$ 3.7 & 37.0 $\pm$ 5.5 \\
\texttt{si\_sd\_replay\_osrm} & 17.7 $\pm$ 2.6 & 99.3 $\pm$ 0.5 & 82.8 $\pm$ 2.5 & 94.8 $\pm$ 0.3 & 35.4 $\pm$ 5.1 \\
\texttt{replay\_merge} & 16.6 $\pm$ 5.8 & 99.9 $\pm$ 0.0 & 84.2 $\pm$ 5.9 & 89.0 $\pm$ 5.4 & 33.1 $\pm$ 11.6 \\
\texttt{si\_replay\_merge} & 15.7 $\pm$ 3.3 & 99.8 $\pm$ 0.2 & 85.0 $\pm$ 3.2 & 86.5 $\pm$ 8.4 & 31.3 $\pm$ 6.5 \\
\texttt{si\_sd\_replay} & 15.4 $\pm$ 0.3 & 99.3 $\pm$ 0.8 & 84.9 $\pm$ 0.8 & 89.7 $\pm$ 0.5 & 30.7 $\pm$ 0.6 \\
\texttt{si\_sd\_replay\_olora} & 15.3 $\pm$ 4.5 & 98.8 $\pm$ 0.3 & 84.5 $\pm$ 4.7 & 79.2 $\pm$ 5.6 & 29.7 $\pm$ 8.3 \\
\texttt{sd\_merge} & 11.8 $\pm$ 1.3 & 99.8 $\pm$ 0.1 & 88.9 $\pm$ 1.3 & 83.4 $\pm$ 4.5 & 23.5 $\pm$ 2.7 \\
\texttt{sd\_replay} & 9.0 $\pm$ 0.5 & 99.6 $\pm$ 0.3 & 91.5 $\pm$ 0.2 & 78.3 $\pm$ 2.8 & 18.0 $\pm$ 1.0 \\
\texttt{si\_replay} & 7.6 $\pm$ 0.8 & 99.9 $\pm$ 0.0 & 93.2 $\pm$ 0.8 & 68.5 $\pm$ 5.6 & 15.3 $\pm$ 1.5 \\
\texttt{si\_sd} & 6.9 $\pm$ 0.6 & 99.3 $\pm$ 0.1 & 93.4 $\pm$ 0.5 & 60.3 $\pm$ 4.9 & 13.7 $\pm$ 1.2 \\
\texttt{si\_sd\_merge} & 6.8 $\pm$ 5.9 & 95.0 $\pm$ 8.0 & 89.3 $\pm$ 1.9 & 53.5 $\pm$ 46.4 & 13.5 $\pm$ 11.7 \\
\texttt{replay} & 4.2 $\pm$ 0.9 & 99.8 $\pm$ 0.1 & 96.6 $\pm$ 1.0 & 41.9 $\pm$ 8.3 & 8.4 $\pm$ 1.8 \\
\texttt{sd} & 3.3 $\pm$ 0.0 & 99.8 $\pm$ 0.1 & 97.5 $\pm$ 0.1 & 33.4 $\pm$ 0.4 & 6.7 $\pm$ 0.1 \\
\texttt{olora} & 1.8 $\pm$ 0.1 & 99.9 $\pm$ 0.1 & 99.1 $\pm$ 0.2 & 18.2 $\pm$ 1.0 & 3.6 $\pm$ 0.2 \\
\texttt{merge} & 1.7 $\pm$ 0.1 & 98.5 $\pm$ 0.6 & 97.8 $\pm$ 0.7 & 17.1 $\pm$ 0.9 & 3.4 $\pm$ 0.2 \\
\texttt{osrm} & 1.6 $\pm$ 0.1 & 98.7 $\pm$ 0.3 & 98.1 $\pm$ 0.3 & 15.5 $\pm$ 1.2 & 3.1 $\pm$ 0.2 \\
\texttt{si\_merge} & 1.5 $\pm$ 0.2 & 97.7 $\pm$ 0.3 & 97.2 $\pm$ 0.5 & 14.6 $\pm$ 1.6 & 2.9 $\pm$ 0.3 \\
\texttt{si} & 1.3 $\pm$ 0.1 & 96.4 $\pm$ 0.2 & 96.1 $\pm$ 0.3 & 12.6 $\pm$ 1.0 & 2.5 $\pm$ 0.2 \\
\texttt{online\_ewc} & 1.0 $\pm$ 0.6 & 86.0 $\pm$ 15.1 & 85.8 $\pm$ 14.6 & 10.3 $\pm$ 6.3 & 2.1 $\pm$ 1.3 \\
\texttt{vanilla} & 1.0 $\pm$ 0.1 & 97.4 $\pm$ 0.4 & 97.4 $\pm$ 0.3 & 10.3 $\pm$ 0.5 & 2.1 $\pm$ 0.1 \\
\hline
\end{tabular}
\caption{All configurations on \symbolqa{}, mean $\pm$ standard deviation over three seeds. $W(k)$ is the final-checkpoint accuracy averaged over the $k$ most recently trained tasks.}
\label{tab:app-all-symbol-qa}
\end{table}

\begin{table}[H]
\centering
\small
\begin{tabular}{lrrrrr}
\hline
configuration & Final & Diag & Forget & $W(10)$ & $W(50)$ \\
\hline
\texttt{si\_replay\_olora} & 42.2 $\pm$ 5.2 & 99.8 $\pm$ 0.0 & 58.2 $\pm$ 5.2 & 93.9 $\pm$ 1.9 & 60.2 $\pm$ 5.7 \\
\texttt{si\_sd\_replay\_merge} & 41.8 $\pm$ 1.4 & 99.3 $\pm$ 0.1 & 58.3 $\pm$ 1.5 & 96.3 $\pm$ 0.5 & 65.5 $\pm$ 0.8 \\
\texttt{si\_replay\_merge} & 41.2 $\pm$ 9.5 & 99.6 $\pm$ 0.2 & 59.1 $\pm$ 9.4 & 98.2 $\pm$ 0.7 & 64.3 $\pm$ 12.1 \\
\texttt{sd\_replay\_merge} & 33.0 $\pm$ 1.8 & 99.4 $\pm$ 0.2 & 67.1 $\pm$ 1.8 & 94.6 $\pm$ 2.7 & 54.9 $\pm$ 1.9 \\
\texttt{replay\_merge} & 32.4 $\pm$ 8.9 & 99.7 $\pm$ 0.1 & 68.0 $\pm$ 9.0 & 97.5 $\pm$ 2.3 & 54.8 $\pm$ 13.7 \\
\texttt{si\_replay\_osrm} & 29.7 $\pm$ 3.0 & 99.5 $\pm$ 0.1 & 70.6 $\pm$ 3.0 & 98.4 $\pm$ 0.5 & 51.2 $\pm$ 5.5 \\
\texttt{si\_sd\_replay} & 18.6 $\pm$ 0.8 & 99.5 $\pm$ 0.1 & 81.7 $\pm$ 0.8 & 86.4 $\pm$ 1.4 & 33.0 $\pm$ 1.7 \\
\texttt{si\_sd\_merge} & 18.3 $\pm$ 0.9 & 99.4 $\pm$ 0.1 & 81.9 $\pm$ 0.9 & 75.7 $\pm$ 2.1 & 33.4 $\pm$ 1.8 \\
\texttt{si\_replay} & 15.5 $\pm$ 2.7 & 99.6 $\pm$ 0.2 & 84.9 $\pm$ 2.6 & 80.4 $\pm$ 6.1 & 27.6 $\pm$ 4.8 \\
\texttt{sd\_merge} & 12.8 $\pm$ 0.8 & 99.6 $\pm$ 0.0 & 87.7 $\pm$ 0.8 & 67.2 $\pm$ 1.5 & 23.4 $\pm$ 1.4 \\
\texttt{sd\_replay} & 9.4 $\pm$ 1.0 & 99.6 $\pm$ 0.1 & 91.2 $\pm$ 1.1 & 62.2 $\pm$ 5.1 & 16.3 $\pm$ 1.6 \\
\texttt{si\_sd} & 8.7 $\pm$ 0.5 & 99.3 $\pm$ 0.1 & 91.5 $\pm$ 0.6 & 58.9 $\pm$ 3.2 & 16.4 $\pm$ 1.1 \\
\texttt{replay} & 7.5 $\pm$ 2.7 & 99.4 $\pm$ 0.2 & 92.9 $\pm$ 2.6 & 52.2 $\pm$ 15.8 & 13.1 $\pm$ 4.8 \\
\texttt{olora} & 5.5 $\pm$ 0.2 & 99.8 $\pm$ 0.1 & 95.2 $\pm$ 0.3 & 30.2 $\pm$ 0.3 & 8.3 $\pm$ 0.1 \\
\texttt{online\_ewc} & 4.5 $\pm$ 0.5 & 89.2 $\pm$ 0.2 & 85.6 $\pm$ 0.8 & 25.7 $\pm$ 0.5 & 7.5 $\pm$ 0.6 \\
\texttt{osrm} & 3.4 $\pm$ 0.7 & 98.6 $\pm$ 0.1 & 96.1 $\pm$ 0.6 & 26.4 $\pm$ 4.7 & 6.2 $\pm$ 1.2 \\
\texttt{merge} & 3.0 $\pm$ 0.3 & 98.7 $\pm$ 0.0 & 96.7 $\pm$ 0.4 & 25.6 $\pm$ 2.1 & 5.6 $\pm$ 0.6 \\
\texttt{sd} & 2.8 $\pm$ 0.0 & 99.7 $\pm$ 0.1 & 97.8 $\pm$ 0.2 & 24.4 $\pm$ 0.3 & 5.2 $\pm$ 0.1 \\
\texttt{si\_merge} & 2.5 $\pm$ 0.0 & 97.5 $\pm$ 0.1 & 95.9 $\pm$ 0.1 & 20.0 $\pm$ 0.8 & 4.5 $\pm$ 0.0 \\
\texttt{si} & 1.8 $\pm$ 0.1 & 95.3 $\pm$ 0.1 & 94.4 $\pm$ 0.2 & 16.0 $\pm$ 0.1 & 3.4 $\pm$ 0.1 \\
\texttt{vanilla} & 1.4 $\pm$ 0.0 & 96.0 $\pm$ 0.7 & 95.6 $\pm$ 0.6 & 12.6 $\pm$ 0.1 & 2.6 $\pm$ 0.0 \\
\hline
\end{tabular}
\caption{All configurations on \llmqa{}, mean $\pm$ standard deviation over three seeds. $W(k)$ is the final-checkpoint accuracy averaged over the $k$ most recently trained tasks.}
\label{tab:app-all-llm-qa}
\end{table}

\begin{table}[H]
\centering
\small
\begin{tabular}{lrrrrr}
\hline
configuration & Final & Diag & Forget & $W(10)$ & $W(50)$ \\
\hline
\texttt{si\_replay\_olora} & 56.4 $\pm$ 13.2 & 98.5 $\pm$ 0.9 & 42.7 $\pm$ 14.1 & 87.5 $\pm$ 3.2 & 70.5 $\pm$ 13.8 \\
\texttt{si\_replay\_merge} & 54.8 $\pm$ 10.4 & 99.9 $\pm$ 0.1 & 45.5 $\pm$ 10.5 & 97.4 $\pm$ 0.5 & 77.9 $\pm$ 9.7 \\
\texttt{replay\_merge} & 48.2 $\pm$ 2.8 & 99.9 $\pm$ 0.0 & 52.3 $\pm$ 2.8 & 98.1 $\pm$ 0.4 & 75.4 $\pm$ 1.9 \\
\texttt{si\_sd\_replay\_merge} & 44.3 $\pm$ 8.7 & 99.5 $\pm$ 0.0 & 55.8 $\pm$ 8.7 & 95.5 $\pm$ 1.9 & 72.7 $\pm$ 11.0 \\
\texttt{si\_replay\_osrm} & 43.2 $\pm$ 2.7 & 99.9 $\pm$ 0.0 & 57.3 $\pm$ 2.8 & 95.6 $\pm$ 0.5 & 70.7 $\pm$ 4.4 \\
\texttt{sd\_replay\_merge} & 39.0 $\pm$ 1.9 & 99.5 $\pm$ 0.2 & 61.1 $\pm$ 2.0 & 96.5 $\pm$ 0.6 & 67.9 $\pm$ 1.9 \\
\texttt{si\_sd\_merge} & 31.6 $\pm$ 0.6 & 99.9 $\pm$ 0.1 & 69.1 $\pm$ 0.5 & 94.3 $\pm$ 0.3 & 57.7 $\pm$ 0.5 \\
\texttt{sd\_merge} & 20.2 $\pm$ 0.4 & 99.9 $\pm$ 0.0 & 80.6 $\pm$ 0.4 & 90.5 $\pm$ 0.6 & 38.8 $\pm$ 0.8 \\
\texttt{si\_replay} & 19.8 $\pm$ 0.9 & 99.9 $\pm$ 0.0 & 81.0 $\pm$ 0.9 & 87.5 $\pm$ 2.4 & 36.5 $\pm$ 1.2 \\
\texttt{si\_sd\_replay} & 15.7 $\pm$ 2.6 & 99.7 $\pm$ 0.0 & 84.9 $\pm$ 2.6 & 83.3 $\pm$ 3.1 & 30.4 $\pm$ 5.0 \\
\texttt{si\_sd} & 14.2 $\pm$ 0.6 & 100.0 $\pm$ 0.0 & 86.6 $\pm$ 0.6 & 81.0 $\pm$ 1.0 & 27.5 $\pm$ 1.1 \\
\texttt{olora} & 13.9 $\pm$ 0.4 & 99.9 $\pm$ 0.0 & 86.9 $\pm$ 0.4 & 46.2 $\pm$ 1.9 & 17.3 $\pm$ 0.3 \\
\texttt{replay} & 12.5 $\pm$ 1.4 & 100.0 $\pm$ 0.0 & 88.3 $\pm$ 1.4 & 78.5 $\pm$ 4.7 & 23.9 $\pm$ 2.7 \\
\texttt{sd\_replay} & 9.5 $\pm$ 0.8 & 99.7 $\pm$ 0.0 & 91.1 $\pm$ 0.8 & 70.1 $\pm$ 2.7 & 18.5 $\pm$ 1.4 \\
\texttt{sd} & 6.6 $\pm$ 0.1 & 100.0 $\pm$ 0.0 & 94.3 $\pm$ 0.1 & 56.1 $\pm$ 1.4 & 12.8 $\pm$ 0.1 \\
\texttt{si\_merge} & 6.5 $\pm$ 0.4 & 99.9 $\pm$ 0.1 & 94.3 $\pm$ 0.4 & 53.2 $\pm$ 4.9 & 12.5 $\pm$ 0.9 \\
\texttt{si} & 4.7 $\pm$ 0.2 & 99.3 $\pm$ 0.1 & 95.6 $\pm$ 0.3 & 40.6 $\pm$ 3.1 & 9.1 $\pm$ 0.6 \\
\texttt{online\_ewc} & 4.5 $\pm$ 0.3 & 99.0 $\pm$ 0.2 & 95.4 $\pm$ 0.2 & 35.7 $\pm$ 4.1 & 8.6 $\pm$ 0.7 \\
\texttt{osrm} & 4.4 $\pm$ 0.4 & 99.8 $\pm$ 0.1 & 96.3 $\pm$ 0.4 & 39.1 $\pm$ 2.7 & 8.5 $\pm$ 0.6 \\
\texttt{merge} & 4.0 $\pm$ 0.4 & 99.8 $\pm$ 0.1 & 96.8 $\pm$ 0.4 & 34.9 $\pm$ 4.4 & 7.7 $\pm$ 0.9 \\
\texttt{vanilla} & 1.3 $\pm$ 0.2 & 97.9 $\pm$ 1.6 & 97.5 $\pm$ 1.5 & 12.9 $\pm$ 1.4 & 2.6 $\pm$ 0.3 \\
\hline
\end{tabular}
\caption{All configurations on \realqa{}, mean $\pm$ standard deviation over three seeds. $W(k)$ is the final-checkpoint accuracy averaged over the $k$ most recently trained tasks.}
\label{tab:app-all-real-qa}
\end{table}

\subsection{Complete factorial tables}
\label{app:res-anova}

\begin{table}[H]
\centering
\small
\begin{tabular}{lrrrr}
\hline
term & effect (percentage) & \% var & $F$ & $p$ \\
\hline
replay & +9.49 & 44.3 & 161.1 & $<10^{-4}$ \\
merge & +5.87 & 16.9 & 61.5 & $<10^{-4}$ \\
sd & +5.66 & 15.8 & 57.3 & $<10^{-4}$ \\
replay $\times$ merge & +3.57 & 6.3 & 22.7 & $<10^{-4}$ \\
si $\times$ merge & -3.04 & 4.6 & 16.6 & 0.000287 \\
si $\times$ sd $\times$ merge & -1.84 & 1.7 & 6.1 & 0.0192 \\
sd $\times$ replay $\times$ merge & -1.31 & 0.8 & 3.0 & 0.0904 \\
si $\times$ replay $\times$ merge & -0.79 & 0.3 & 1.1 & 0.297 \\
si $\times$ replay & +0.72 & 0.3 & 0.9 & 0.341 \\
sd $\times$ merge & +0.56 & 0.2 & 0.6 & 0.462 \\
si & +0.34 & 0.1 & 0.2 & 0.649 \\
si $\times$ sd & -0.28 & 0.0 & 0.1 & 0.712 \\
si $\times$ sd $\times$ replay $\times$ merge & +0.17 & 0.0 & 0.1 & 0.822 \\
sd $\times$ replay & -0.16 & 0.0 & 0.0 & 0.837 \\
si $\times$ sd $\times$ replay & +0.09 & 0.0 & 0.0 & 0.902 \\
\hline
\end{tabular}
\caption{Complete saturated $2^4$ factorial on \symbolqa{}, $N=48$, error df 32, RMSE 2.59, 91\% of variance explained. The standard error of any effect is 0.75 percentage.}
\label{tab:app-anova-symbol-qa}
\end{table}

\begin{table}[H]
\centering
\small
\begin{tabular}{lrrrr}
\hline
term & effect (percentage) & \% var & $F$ & $p$ \\
\hline
replay & +18.49 & 43.9 & 341.8 & $<10^{-4}$ \\
merge & +14.91 & 28.6 & 222.3 & $<10^{-4}$ \\
replay $\times$ merge & +9.44 & 11.4 & 89.0 & $<10^{-4}$ \\
si & +5.76 & 4.3 & 33.2 & $<10^{-4}$ \\
sd & +5.02 & 3.2 & 25.2 & $<10^{-4}$ \\
sd $\times$ replay & -3.45 & 1.5 & 11.9 & 0.00159 \\
si $\times$ replay & +2.93 & 1.1 & 8.6 & 0.00624 \\
sd $\times$ replay $\times$ merge & -2.62 & 0.9 & 6.9 & 0.0134 \\
sd $\times$ merge & +1.69 & 0.4 & 2.9 & 0.101 \\
si $\times$ sd & +1.59 & 0.3 & 2.5 & 0.121 \\
si $\times$ sd $\times$ replay & -1.29 & 0.2 & 1.7 & 0.205 \\
si $\times$ sd $\times$ replay $\times$ merge & -0.23 & 0.0 & 0.1 & 0.823 \\
si $\times$ replay $\times$ merge & +0.20 & 0.0 & 0.0 & 0.845 \\
si $\times$ merge & -0.13 & 0.0 & 0.0 & 0.899 \\
si $\times$ sd $\times$ merge & -0.10 & 0.0 & 0.0 & 0.919 \\
\hline
\end{tabular}
\caption{Complete saturated $2^4$ factorial on \llmqa{}, $N=48$, error df 32, RMSE 3.46, 96\% of variance explained. The standard error of any effect is 1.00.}
\label{tab:app-anova-llm-qa}
\end{table}

\begin{table}[H]
\centering
\small
\begin{tabular}{lrrrr}
\hline
term & effect (percentage) & \% var & $F$ & $p$ \\
\hline
merge & +20.54 & 36.4 & 393.4 & $<10^{-4}$ \\
replay & +19.34 & 32.3 & 348.9 & $<10^{-4}$ \\
replay $\times$ merge & +11.68 & 11.8 & 127.1 & $<10^{-4}$ \\
sd $\times$ replay & -10.32 & 9.2 & 99.2 & $<10^{-4}$ \\
si & +6.30 & 3.4 & 37.0 & $<10^{-4}$ \\
sd $\times$ replay $\times$ merge & -4.87 & 2.0 & 22.1 & $<10^{-4}$ \\
sd & +3.66 & 1.2 & 12.5 & 0.00128 \\
si $\times$ sd $\times$ replay & -1.93 & 0.3 & 3.5 & 0.0711 \\
sd $\times$ merge & +1.74 & 0.3 & 2.8 & 0.103 \\
si $\times$ sd & +1.33 & 0.2 & 1.6 & 0.208 \\
si $\times$ sd $\times$ replay $\times$ merge & -0.62 & 0.0 & 0.4 & 0.553 \\
si $\times$ replay $\times$ merge & -0.57 & 0.0 & 0.3 & 0.588 \\
si $\times$ sd $\times$ merge & +0.54 & 0.0 & 0.3 & 0.605 \\
si $\times$ merge & +0.17 & 0.0 & 0.0 & 0.868 \\
si $\times$ replay & +0.06 & 0.0 & 0.0 & 0.95 \\
\hline
\end{tabular}
\caption{Complete saturated $2^4$ factorial on \realqa{}, $N=48$, error df 32, RMSE 3.59, 97\% of variance explained. The standard error of any effect is 1.04.}
\label{tab:app-anova-real-qa}
\end{table}

\subsection{Greedy composition paths}
\label{app:res-buildup}

\begin{figure}[H]
\centering
\includegraphics[width=\linewidth]{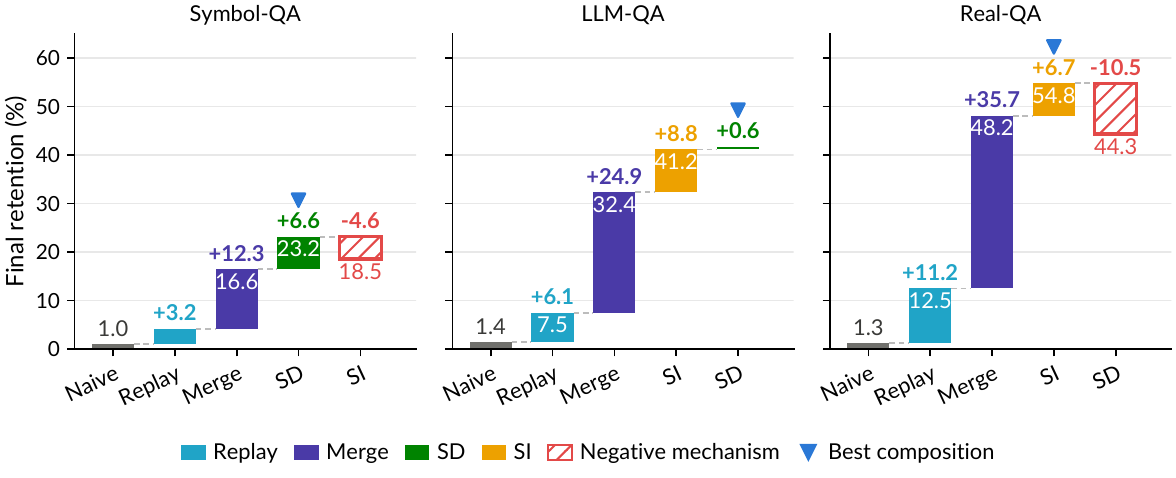}
\caption{Greedy buildup paths through the $2^4$ factorial. Each step adds the remaining mechanism with the highest final retention under the current composition. Triangles mark the strongest composition along each path.}
\label{fig:app-buildup}
\end{figure}

Figure~\ref{fig:app-buildup} presents a greedy nested path through the factorial for each dataset. Starting from naive fine-tuning, each step adds the remaining mechanism that gives the highest final retention when combined with the mechanisms already selected. Every method is trained independently, so the path summarizes comparisons within the factorial rather than a sequence of training stages.

\subsection{Retention matrices}
\label{app:res-matrices}

\begin{figure}[H]
\centering
\includegraphics[width=\linewidth]{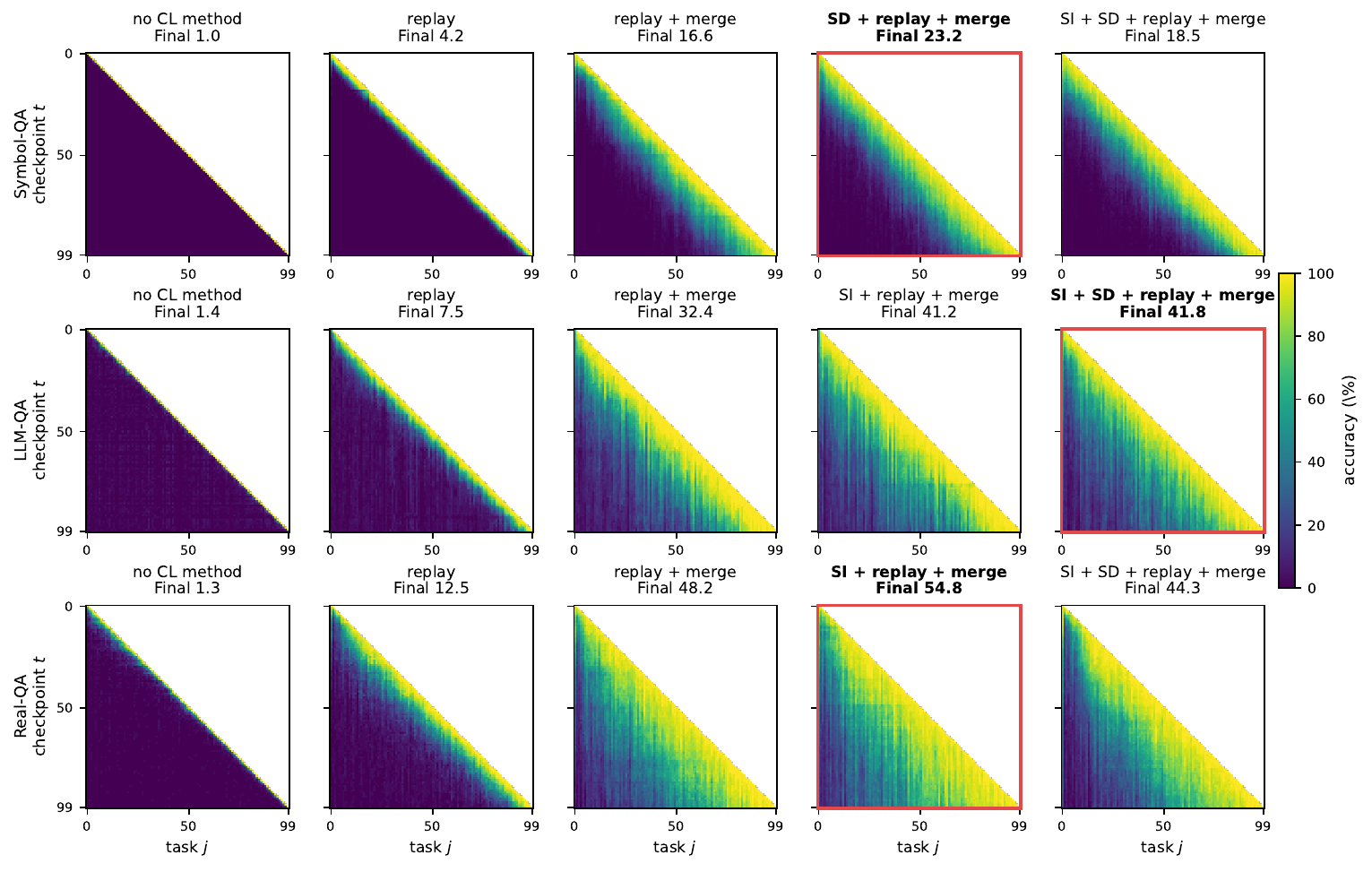}
\caption{Temporal accuracy matrices $M_{i,j}$ along the composition chain, one row per dataset. Each column adds one mechanism and never removes one. The panel outlined in red is the strongest configuration in that row. Panels show the median seed of three, so no panel is a favorable draw.}
\label{fig:matrices}
\end{figure}

Figure~\ref{fig:matrices} shows the retention matrix from which all reported metrics are computed. Each panel is one temporal accuracy matrix $M_{i,j}$: row $i$ is the checkpoint after task $i$, column $j$ is the task being evaluated, and only the lower triangle is observable because a task cannot be evaluated before it is trained. The diagonal represents the performance on the current training task, and everything below it is what remains of a memory as later tasks arrive. Read Figure~\ref{fig:matrices} left to right, the bright band below the diagonal widens as mechanisms are added: naive fine-tuning retains only a narrow strip along the diagonal, and the leading composition keeps a broad region alive for tens of tasks. Read down a column, the same recipe covers progressively more of the triangle as the data becomes more natural. Two features of the main text are directly visible here. The band has a soft outer edge rather than a hard boundary, which is the graded decay that the survival curves quantify. The configurations form a nested sequence, but adding another anchor does not always improve retention. Adding the weight anchor lowers retention on \symbolqa{}, while adding the function anchor lowers retention on \realqa{}.

\subsection{Task-Growing Low-Rank Allocation Experiments}
\label{app:res-allocation}

The factorial compares shared LoRA and merged LoRA, two low-rank allocation rules that keep learner state independent of the horizon. To compare them with task-growing low-rank allocation rules, each dataset also uses its winning composition with merged LoRA replaced by OSRM or O-LoRA, holding every anchor and hyperparameter fixed. These cells are the only ones in the study whose retained state size grows with the number of tasks, and they are reported separately for that reason.

Additional task-growing state does not provide a consistent retention gain. Relative to the matched merged LoRA composition, O-LoRA changes final retention by $-3.3$, $+1.0$, and $+1.6$ points on \symbolqa{}, \llmqa{}, and \realqa{}. Sequential OSRM changes it by $-0.8$, $-11.5$, and $-11.6$ points. O-LoRA gives the highest mean retention on the two natural-language datasets, but its advantage over the best bounded composition is only 0.4 and 1.6 points, which is within seed variation. Section~\ref{app:res-capability} shows that the low-rank allocation rule has a larger effect on general capability.

\begin{table}[H]
\centering
\small
\begin{tabular}{lrrr}
\hline
$O(T)$ low-rank allocation & \symbolqa{} & \llmqa{} & \realqa{} \\
\hline
O-LoRA & $-3.3\%$ & $+1.0\%$ & $+1.6\%$ \\
Sequential OSRM & $-0.8\%$ & $-11.5\%$ & $-11.6\%$ \\
\hline
\end{tabular}
\caption{Change in final retention after replacing merged LoRA with O-LoRA or sequential OSRM in the composition selected for each dataset.}
\label{tab:allocation-swap}
\end{table}

\subsection{Compositions that fail the acquisition gate}
\label{app:res-failures}

Two cells fail to memorize rather than fail to retain, and both compose a weight anchor with the fold: \verb|si_sd_merge| reaches
$\finalacc = 6.8 \pm 5.9$ with $\diagacc = 95.0 \pm 8.0$, and
\verb|online_ewc| reaches $1.0 \pm 0.6$ with $\diagacc = 86.0 \pm 15.1$. The large standard deviations are the signature: on some seeds the accuracy-matrix
diagonal collapses outright while on others training proceeds normally, so the mean describes a mixture of two behaviours rather than a typical run. Every
other configuration in the study holds $\diagacc$ between 96 and 100.

This is the failure mode the compatibility rule predicts. A quadratic penalty is defined in the coordinates in which its importance estimates were measured; merged LoRA folds the adapter into the base weights and re-initializes,
so those coordinates no longer denote the same function, and the penalty resists motion the model needs in order to fit the current task. The consequence is a loss of plasticity, which is why it appears on the diagonal rather than in the off-diagonal decay.

\subsection{Seed variance and the generation-temperature control}
\label{app:res-seeds}

Seed standard deviations partition cleanly by whether a configuration carries the data anchor. Replay-bearing stacks give 2 to 6 points
(\verb|replay_merge| $\pm 5.8$, \verb|sd_replay_merge| $\pm 4.2$), while configurations without replay are tight (\verb|sd_merge| $\pm 1.3$, \verb|sd| $\pm 0.0$, singletons $\pm 0.1$ to $\pm 0.2$).

A controlled comparison isolates the mechanism. Re-running one composition at replay generation temperature 0.7 rather than 1.5, with three seeds and all else fixed, gives $21.2 \pm 7.9$ against $20.8 \pm 2.1$. The means are
statistically indistinguishable while the seed spread differs roughly fourfold, so generation diversity acts on the variance of the outcome and not on its expectation. Sharp, low-entropy replay concentrates the pseudo-data on few modes, and whether those modes align with a given seed's trajectory determines the run. This is also why single-seed probes are unsafe here: the low-temperature arm contains a seed reaching 30.3 alongside seeds at 16.2 and
17.1, and reporting the first alone would have suggested a large improvement where there is none.

Two byproducts of that re-run are worth recording. The outlying seed reproduced the same value across a three-week interval of repository changes, which is direct evidence that training is deterministic given seed and hyperparameters. And the same winner's-curse mechanism that operates over configurations (Section~\ref{app:res-selection}) operates over hyperparameters.

\subsection{Held-out general capability evaluation}
\label{app:res-capability}

We evaluate the final checkpoints and the unmodified base model on GSM8K~\citep{cobbe2021training}, MATH~\citep{hendrycks2021measuring}, MGSM~\citep{shi2022language}, and MMLU-Redux~\citep{gema2025we}. The base model scores 83.6\%, 57.6\%, 67.5\%, and 68.7\%, respectively, with an unweighted average of 69.4\%. We average trained-model scores over three seeds. GSM8K, MATH, and MGSM are scored by extracting and checking answers from generated text.

For MMLU-Redux, we select an answer directly from the model's probabilities rather than generating text. Each prompt contains two fixed examples followed by the question and four choices labeled A through D, and ends with \texttt{Answer:}. We compare the next-token log probabilities for the four answer labels and select the label with the highest score. The evaluator uses the first token ID obtained by tokenizing each letter with a preceding space. This \emph{letter log-likelihood} score is the logarithm of the probability assigned to a candidate answer-label token, not to the full choice text. Every question receives a prediction without requiring the model to generate a valid answer letter. The scoring therefore avoids answer-extraction failures, although performance can still depend on the prompt.

\begin{figure}[H]
\centering
\includegraphics[width=\linewidth]{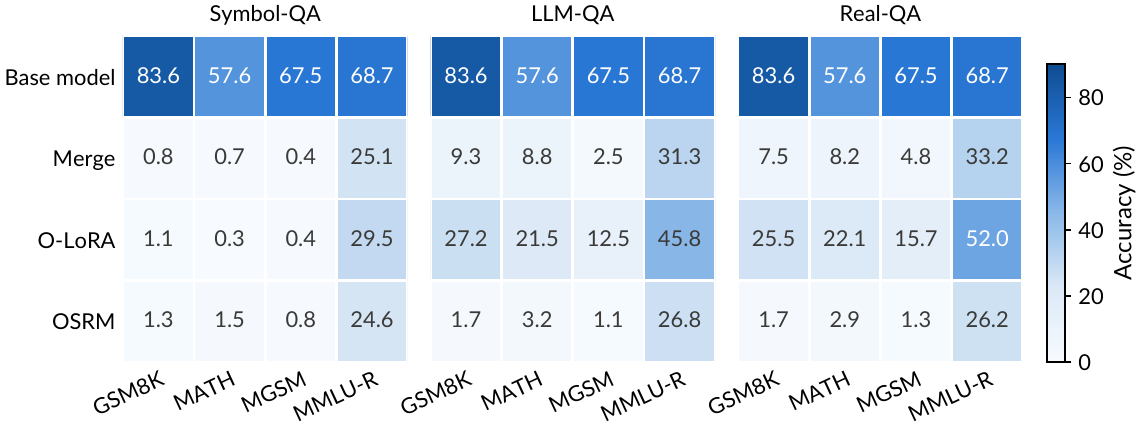}
\caption{General capability after 100 tasks, averaged over three seeds. Anchors and hyperparameters from TSH are fixed across allocation rules: all three anchors on \symbolqa{}, and data and weight anchors on \llmqa{} and \realqa{}. O-LoRA better preserves general capability on the two natural-language datasets.}
\label{fig:capability}
\end{figure}

Figure~\ref{fig:capability} compares general capability across low-rank allocation rules. O-LoRA changes final retention only slightly relative to merged LoRA, but the difference in general capability is larger on the two natural-language datasets. Averaged across the four benchmarks, O-LoRA achieves 26.8\% accuracy after \llmqa{} and 28.8\% after \realqa{}. Merged LoRA reaches 13.0\% and 13.4\%, while sequential OSRM reaches 8.2\% and 8.0\%. The O-LoRA compositions also achieve 45.8\% and 52.0\% on MMLU-Redux, respectively. Even so, their four-benchmark averages remain more than 40 percentage points below the base model.

After \symbolqa{}, all three allocation rules in Figure~\ref{fig:capability} reduce average general capability accuracy to below 8.0\%. Some GSM8K generations repeat fragments of the training format instead of providing mathematical answers. However, output generation failures do not fully explain the low scores. MMLU-Redux accuracy remains near the 25\% chance level even when answer labels are selected directly from their log probabilities.

O-LoRA leaves the base parameters unchanged and accumulates one adapter per task. Removing these adapters restores the original model, but also removes the learned task updates. Merged LoRA and sequential OSRM instead incorporate each update into the base parameters, so the original model cannot be recovered simply by detaching an adapter. This recoverability is distinct from preserving general capability with the learned updates active. Our comparison does not isolate whether this structural difference causes O-LoRA's higher general capability scores. Both O-LoRA and sequential OSRM retain state that grows linearly with the number of tasks, and neither prevents substantial general capability loss after 100 tasks.

\subsection{Selection versus evaluation}
\label{app:res-selection}

The search of Section~\ref{sec:search} and the factorial of Section~\ref{sec:results} disagree about which configuration is best on \symbolqa{}: the search selects \verb|si_sd_replay_merge| and the final evaluation ranks \verb|sd_replay_merge| above it. The two were separated by 1.0 point on the development order, which is inside the seed spread of either.

This is the expected behaviour of an argmax over many noisy candidates. The selected configuration regresses from 23.4 to 18.5 on the report order, while the eventual leader is stable at 22.4 and 23.2. Because \symbolqa{} tasks are exchangeable by construction, the development and default orders are draws from the same distribution, so the movement is seed noise rather than overfitting to a particular ordering. The design anticipates this: the search selects a family under a budget, and the factorial provides unbiased estimates and the decomposition. It is also why the paper reports a recipe and a set of
effects rather than crowning a single configuration.

\subsection{Controls}
\label{app:res-controls}

Forward transfer is $0.0 \pm 0.0$ on \symbolqa{} and below 3.1 points elsewhere, as expected when a task is unseen before its turn; the small positive values on the natural-language datasets reflect shared surface form rather than knowledge of the specific facts. The \verb|\boxed{}| emission rate is 100\% on every configuration and both at the diagonal and at the final checkpoint, so no configuration loses measured accuracy through format failure, and content retention is separated cleanly from format retention.

\end{document}